\documentclass[10pt,twocolumn,letterpaper]{article}

\usepackage{iccv}              

\definecolor{iccvblue}{rgb}{0.21,0.49,0.74}
\usepackage[pagebackref,breaklinks,colorlinks,allcolors=iccvblue]{hyperref}
\usepackage{multirow}
\usepackage{makecell}
\newtheorem{theorem}{Theorem}[section]

\usepackage{algorithm}
\usepackage{algorithmic}
\usepackage[accsupp]{axessibility}

\usepackage{pifont}
\usepackage{indentfirst} 
\def\paperID{10853	
} 
\def\confName{ICCV}
\def\confYear{2025}
\title{FREE-Merging: Fourier Transform for Efficient Model Merging }

\author{Shenghe Zheng\\
Harbin Institute of Technology\\
{\tt\small shenghez.zheng@gmail.com}
\and
Hongzhi Wang\thanks{Corresponding Author}\\
Harbin Institute of Technology\\
{\tt\small wangzh@hit.edu.cn }
}

\begin{document}
\maketitle
\begin{abstract}

With the rapid growth of deep learning, there is an increasing availability of open-source models for various tasks. However, single fine-tuned models often fall short of meeting the diverse needs of users. Model merging has thus emerged as an efficient method to integrate the capabilities of existing models into a unified model. Nevertheless, existing model merging methods face challenging trade-offs between performance and deployment costs, primarily due to task interference. For the first time, we reveal that task interference is evident in the frequency domain of model parameters, yet current efforts only focus on spatial domain solutions, which are largely ineffective in addressing frequency domain interference. 
To mitigate the impact of frequency domain interference, we propose \textbf{FR-Merging}, an innovative method that effectively filters harmful frequency domain interference on the backbone with minimal computational overhead. Since performance loss is inevitable with cost-free methods, we propose a lightweight task-specific expert module that dynamically compensates for information loss during merging. This proposed framework, \textbf{FREE-Merging} (FR-Merging with experts), strikes a balanced trade-off between training cost, inference latency, storage requirements, and performance. We demonstrate the effectiveness of both FR-Merging and FREE-Merging on multiple tasks across CV, NLP, and Multi-Modal domains and show that they can be flexibly adapted to specific needs. Our code is available at:  \href{https://github.com/Zhengsh123/FREE-Merging}{https://github.com/Zhengsh123/FREE-Merging}.

\end{abstract}    
\section{Introduction}
\label{sec:intro}

In the current era of deep learning, the pretrain-finetune paradigm is a standard procedure~\cite{radford2021learning,devlin-etal-2019-bert,touvron2023llama}. However, edge applications often lack powerful computing resources and data, favoring ready-to-use models for specific scenarios. The recent growth of open-source platforms like HuggingFace~\cite{wolf2019huggingface} has made this feasible. Yet, practical demands are often not fully covered by a single existing model and are typically a union of several models~\cite{yangadamerging,jin2022dataless}. However, running separate models for each subtask increases storage and inference costs, especially with large models. Model merging attempts to combine existing models to build a model capable of addressing all target tasks~\cite{ilharco2022editing,matena2022merging,yadav2024ties}. This method reduces training costs, data privacy issues in multi-task learning (MTL)~\cite{zhang2021survey, huang2024emr} and deployment costs, garnering widespread attention.

While model merging aims to reduce training costs, it often suffers from performance loss due to task interference~\cite{huang2024emr}. To mitigate this issue, existing methods develop along three directions. The first direction computes merging coefficients based on weights or data~\cite{jin2022dataless,matena2022merging}. The second uses the parameter differences before and after fine-tuning~(\textit{i.e.} task vectors) for merging~\cite{ilharco2022editing,yadav2024ties,yangadamerging,zhou2024metagpt}. The third method introduces independent knowledge for each task~\cite{huang2024emr,lu2024twin}.  While the third way balances performance and costs, it still faces two critical limitations. First, it neglects backbone optimization during merging, resulting in subpar performance. Second, it requires high computation and storage, making it impractical for edge deployment.

\begin{figure*}  
\centering  
\includegraphics[width=1 \textwidth]{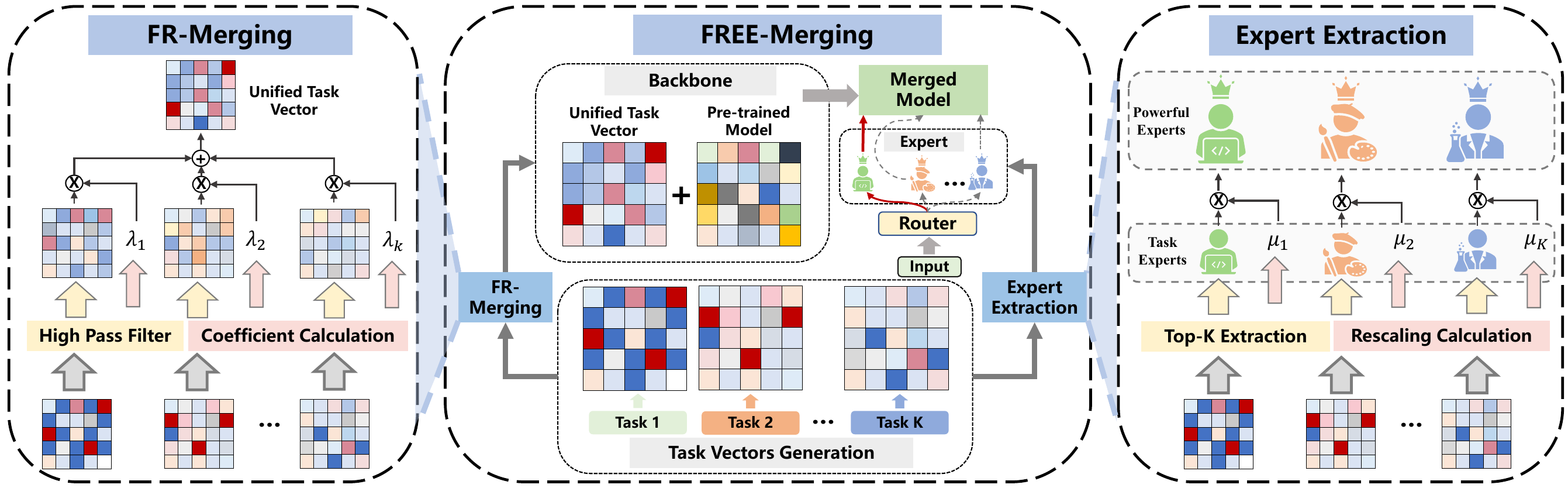} 
\vspace{-0.4cm}
\caption{The workflow of FREE-Merging involves two main steps. First, FR-Merging utilizes high-pass filtering to remove harmful specialized information from each model, thereby constructing a high-quality merged backbone. Second, it employs lightweight extraction of task experts, which are dynamically added during inference to mitigate the impact of task interference.} 
\label{figure:workflow.}  
\end{figure*}

We argue that an effective model merging method requires addressing both of the above issues. First, since the backbone constitutes the majority of parameters and thus determines performance, an efficient backbone merging method is essential to reduce performance loss. Then, given the \textit{no free lunch} theorem, cost-free merging cannot avoid task interference. Therefore, introducing task-specific experts is necessary. However, they must remain lightweight to avoid storage and inference overhead. Thus,  we propose FREE-Merging, a two-stage method that ensures an efficient backbone and lightweight experts as shown in Fig~\ref{figure:workflow.}. 

For backbone merging, the key lies in eliminating task interference~\cite{yangrepresentation}. We identify a compelling phenomenon in which task interference is evident in the frequency domain of model parameters. And we present in Sec.~\ref{sec:Task_Conflict} that frequency-domain interference is strongly correlated with performance. Notably, existing methods, which focus only on spatial-domain operations, fail to address frequency-domain interference, leading to suboptimal outcomes. Experimental evidence reveals that task interference is severe in the low-frequency region. This occurs because, high-frequency signals represent fine-grained variations, while low-frequency parts capture the global structure~\cite{gonzales1987digital}, which is more likely to contain task-specific information that leads to task interference. Leveraging the inherent redundancy in fine-tuned parameters~\cite{yu2024language,liu2024bitdelta,xurandom}, we propose to directly filter out low-frequency parts with severe task interference. As detailed in Sec.~\ref{sec:fr-merging}, this high-pass filtering substantially enhances generalization with minimal performance loss, preventing task interference during merging. Our lightweight yet highly effective approach, FR-Merging, is the first to apply Fourier filtering to neural network parameters, showing its potential for broader applications.

After obtaining a high-performance backbone through FR-Merging, we use a lightweight task-specific expert to recover the inevitable information loss with minimal storage.  Unlike existing methods that store extensive knowledge, we rescale low-frequency signals, which store task-specific information, retaining task expertise with only 1\% of the parameters. Inspired by MoE~\cite{shazeer2016outrageously,zhou2022mixture}, we employ a router for dynamic expert routing, enhancing model flexibility.

In summary, we introduce FREE-Merging, which utilizes FR-Merging to reduce task interference and incorporates lightweight experts to eliminate information loss of model merging. Our main contributions are: 1). We identify frequency-domain task interference as a critical factor that greatly impacts merging performance but remains challenging to address through spatial-domain methods. 2). To tackle frequency-domain interference, we propose FR-Merging, which reduces interference while preserving performance, constructing the merged backbone. 3). We theoretically validate the necessity of introducing new information during merging and propose an effective expert construction method. 4). Extensive experiments show that FR-Merging and FREE-Merging effectively enhance performance across vision, language, and multi-modal tasks.

\section{Related Work}
\label{sec:related work}
\noindent\textbf{Model Merging.}
Model merging integrates existing models to handle multiple tasks without training~\cite{ilharco2022editing,jin2022dataless,yang2024model}, but faces task interference challenges~\cite{yadav2024ties,qi2024less}. Simple averaging~\cite{wortsman2022model} causes great performance degradation. Methods like Fisher-Merging~\cite{matena2022merging} and RegMean~\cite{jin2022dataless} use matrices to determine merging coefficients but are computationally expensive or data-intensive, limiting edge deployment. Task Arithmetic~\cite{ilharco2022editing} merges task vectors instead of weights, while Ties-Merging~\cite{yadav2024ties} resolves parameter conflicts, and AdaMerging~\cite{yangadamerging} automates coefficient selection.
However, these approaches have data distribution requirements and limited applicability. Techniques such as DARE~\cite{yu2024language} and PCB-Merging~\cite{du2024parameter} mitigate performance drops but still face task conflicts. EMR-Merging~\cite{huang2024emr} and Twin-Merging~\cite{lu2024twin} store task-specific knowledge but require large storage and neglect backbone optimization. Our method uses high-pass filtering and lightweight experts to balance merging costs and effectiveness across various tasks. For further details, please refer to Appendix~\ref{appendix:baselines}.

\noindent \textbf{Model Ensemble.} Model ensemble combines outputs from multiple models~\cite{ganaie2022ensemble,dong2020survey,jiang2023llm}, but with large models, it faces storage and inference challenges. In contrast, Model merging enables a single model to solve multiple tasks, significantly reducing storage and inference costs.

\noindent \textbf{Multi-task Learning.} Multi-task learning~(MTL) aims to solve multiple tasks using a single model~\cite{chen2024multi,vandenhende2021multi,liu2023hierarchical}, but in the era of large models, it faces challenges like high training costs, data privacy issues, and expertise requirements~\cite{yangadamerging}. In contrast, model merging uses existing open-source models to create a multi-task model with little or no training, significantly reducing deployment costs.

\section{Preliminaries}
\label{sec:Preliminaries}
\noindent\textbf{Notation.} Assume that $f_{\theta}(x)\rightarrow y$ represents a neural network parameterized by $\theta$, where $x \in R^m$ is the input and $y \in R^n$ is the output. Recent studies have focused on model merging via task vectors. For task $k$, the task vector is defined as $v_k =\theta_k-\theta_{pre}$, where $\theta_{k}$ is the fine-tuned model for task $k$, and $\theta_{pre}$ is the corresponding pre-trained model. Additionally, $e_k$ represents the task expert that retains the ability to solve the task $k$ with few parameters.

\noindent \textbf{Problem Definition.} The objective of model merging is to combine models $\{\theta_k\}_{k=1}^K$ for tasks $T=\{t_i\}_{i=1}^K$ into a single model $\theta_m$ that can handle all tasks in $T$. We focus on a practical scenario where a shared pre-trained model $\theta_{pre}$ serves as the base for all fine-tuned models. It is formulated as $\theta_m = \lambda \sum_{k=1}^K \hat G(\theta_k)$, and with task vector, it becomes $\theta_m=\theta_{pre}+\lambda \Sigma_{k=1}^K G(v_k)$, where $\lambda$  is the coefficient, and $\hat G/G$ is transformation on the parameters. Assuming a shared pre-trained model aligns with trends toward foundation models, like LLaMa~\cite{touvron2023llama} for language and CLIP~\cite{radford2021learning} for vision, which are then fine-tuned for specific tasks. 
\section{Task Interference in Frequency Domain}
\label{sec:Task_Conflict}
\begin{figure}[t] 
\centering  
\includegraphics[width=1.0 \linewidth]{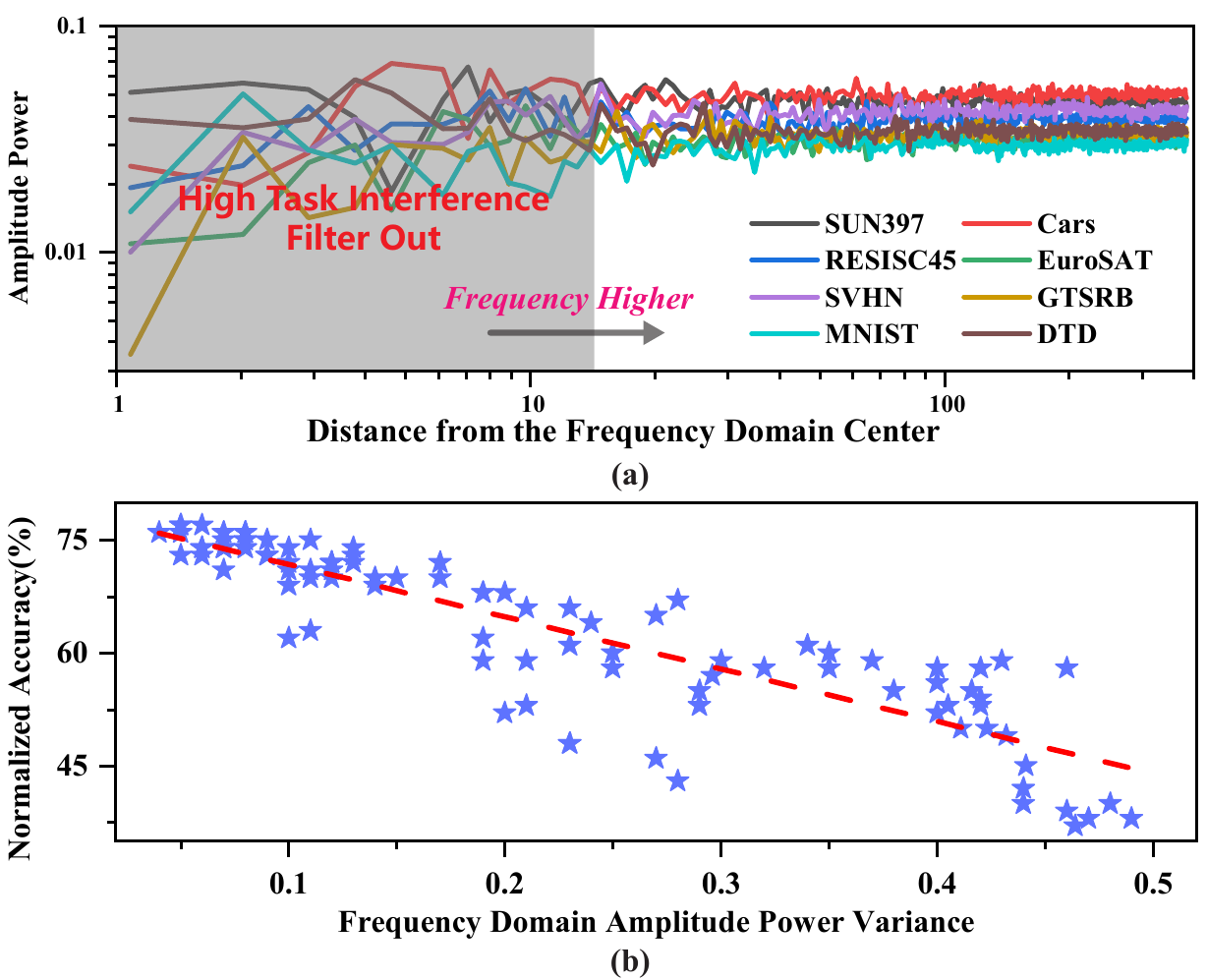} 
\vspace{-0.5cm}
\caption{a) The frequency domain amplitude power distribution of an attention head across different fine-tuned ViT-B-32 models. b) The correlation between the merging performance of ViT-B-32 and the variance of frequency domain amplitude power. Normalized accuracy is defined as the ratio of merged model performance to fine-tuned model performance on the same task.} 
\label{figure:freq_vit.}  
\end{figure} 

Task interference is a key factor that leads to the decrease in merging performance. While current methods focus on addressing interference in the spatial domain~\cite{yadav2024ties,yang2024representation,yu2024language}, we observe that task vectors from different fine-tuned models also exhibit great misalignment in the frequency domain.

As shown in Fig.~\ref{figure:freq_vit.}(a), we visualize the amplitude power distribution in the frequency domain of task vectors from a ViT-B-32 attention head fine-tuned on various tasks~\cite{ilharco2022editing}. Here, greater distance from the center represents a higher frequency. Since amplitude indicates signal strength, merging two signals with vastly different strengths inevitably results in the weaker signal being masked. Therefore, task interference in the low-frequency part are severe.

Furthermore, in Fig.\ref{figure:freq_vit.}(b), we train multiple groups of models on eight image tasks~\cite{ilharco2022editing} and find a clear negative correlation between the variance of amplitude power among models and the normalized merging performance (details can be found in Appendix~\ref{appendix:task_inteference}), confirming the great impact of frequency-domain interference on merging performance.

\begin{table}[t]
    \centering
    \caption{The impact of current methods on the mean variance of frequency-domain amplitude across 8 image classification tasks.}\label{tab:cur_var}
    \scalebox{0.9}{
    \begin{tabular}{ccc}
    \hline
        Method & ViT-B/32 & ViT-L/14 \\ \hline
        Task Arithmetic~\cite{ilharco2022editing} & 0.059   &0.110 \\ DARE~\cite{yu2024language} &0.057($\downarrow $3\%)   &0.108($\downarrow $2\%)  \\ 
        Ties-Merging~\cite{yadav2024ties} &0.058($\downarrow $2\%)    &0.109($\downarrow $1\%)  \\ Breadcrumbs~\cite{davari2023model} &0.058($\downarrow $2\%) &0.107($\downarrow $3\%) \\
        PCB-Merging~\cite{du2024parameter}&0.056($\downarrow $5\%)  &0.107($\downarrow $3\%) \\
        \hline
        FR-Merging(ours)& \textbf{0.045($\downarrow $24\%)}  &\textbf{0.088($\downarrow $20\%)} \\
        \hline
    \end{tabular}
    }
    
\end{table}

Additionally, as shown in Table~\ref{tab:cur_var}, we show that existing works addressing task interference in the spatial domain fail to effectively mitigate frequency-domain task interference. Therefore, we propose our approach to tackle this issue.
\section{Methodology}\label{sec:methodology}
This section presents FREE-Merging: Sec.~\ref{sec:motivation} explains the need for a two-part design, Sec.~\ref{sec:fr-merging} and Sec.~\ref{sec:expert} cover the FR-Merging for the backbone and the lightweight expert module, respectively, followed by the FREE-Merging in Sec.~\ref{sec:free_merging} and a complexity analysis in Sec.~\ref{sec:method:analysis}.

\subsection{Motivation}\label{sec:motivation}
We first analyze why both merging optimization for the backbone and the introduction of experts are necessary.

First, backbone optimization is crucial but overlooked by current expert-based merging ways. As the backbone holds most parameters, its performance is key to the overall success. For instance, upgrading from LLaMa2~\cite{touvron2023llama} to LLaMa3~\cite{dubey2024llama} greatly improves results even with the identical fine-tuning. Thus, we propose a training-free merging method to reduce task interference in the backbone.

Next, we present a theorem showing the necessity of task experts. Based on Theorem~\ref{theorem:moe need}, introducing additional information is essential. For proof, see Appendix~\ref{appendix:proofs}, and the effectiveness of task experts is proved in Appendix~\ref{appendix:dis:task experts}.

\begin{theorem}\label{theorem:moe need} 
\textbf{Model Merging has no free lunch.}
Merged model $\theta_m$ cannot simultaneously retain the capabilities of $\{\theta_k\}_{k=1}^K$ without introducing additional information.
\end{theorem}

\subsection{FR-Merging}\label{sec:fr-merging}

We propose an effective backbone merging method by addressing task interference in the low-frequency fine-tuned parameters as discussed in Sec.~\ref{sec:Task_Conflict}, which is hard to resolve in the spatial domain. Given the high redundancy of fine-tuned parameters~\cite{yu2024language}, we apply high-pass filtering to directly remove the interfering regions. Let $v(x,y)$ be the task vector and $G(x,y)$ the transformed result. Then we have:  
\begin{equation}\label{eq: fourier}
G(x, y) = \mathcal{F}^{-1} \{ H(\eta, \gamma) \cdot \mathcal{F}\{ v(x, y) \} \},
\end{equation}
where
\begin{align}\label{eq: fourier 2}
H(\eta, \gamma) &= \begin{cases} 
      1,  \quad \sqrt{\eta^2 + \gamma^2} \geq D_0 \\
      0,  \quad \sqrt{\eta^2 + \gamma^2} < D_0 
      \end{cases}, 
\end{align}
where $\mathcal{F}$ and $\mathcal{F}^{-1}$ represent the Fourier and inverse Fourier transform, respectively. $D_0$ is the adjustable cutoff frequency. And $\eta$ and $\gamma$ are the distances from the frequency domain center along the $x$- and $y$-axes.

Next, we analyze why high-pass filtering is effective from both intuitive and experimental perspectives. Fine-tuned weights occupy distinct positions in the loss landscape, and linear interpolation often leads to high-loss regions, causing task interference~\cite{ainsworthgit,yang2024representation}. To address this, we minimize model differences while preserving performance, bringing them closer in the loss landscape to increase the likelihood of the merged model falling into a loss basin~\cite{wu2023pi}.

Results in Sec.~\ref{sec:Task_Conflict} show great differences in low-frequency regions. This is because, if we consider neural network parameters as signals carrying training information, inspired by image processing~\cite{nguyen2022fourierformer,buchholz2022fourier}, high-frequency signals represent sharp variations, while low-frequency parts define the overall structure~\cite{chu2023rethinking}. Since most information resides in the structure, filtering out low-frequency signals removes key task-specific information. Then, the likelihood that the interpolation falls within the loss basin increases~\cite{ainsworthgit}.

\begin{figure}[t] 
\centering  
\includegraphics[width=0.9 \linewidth]{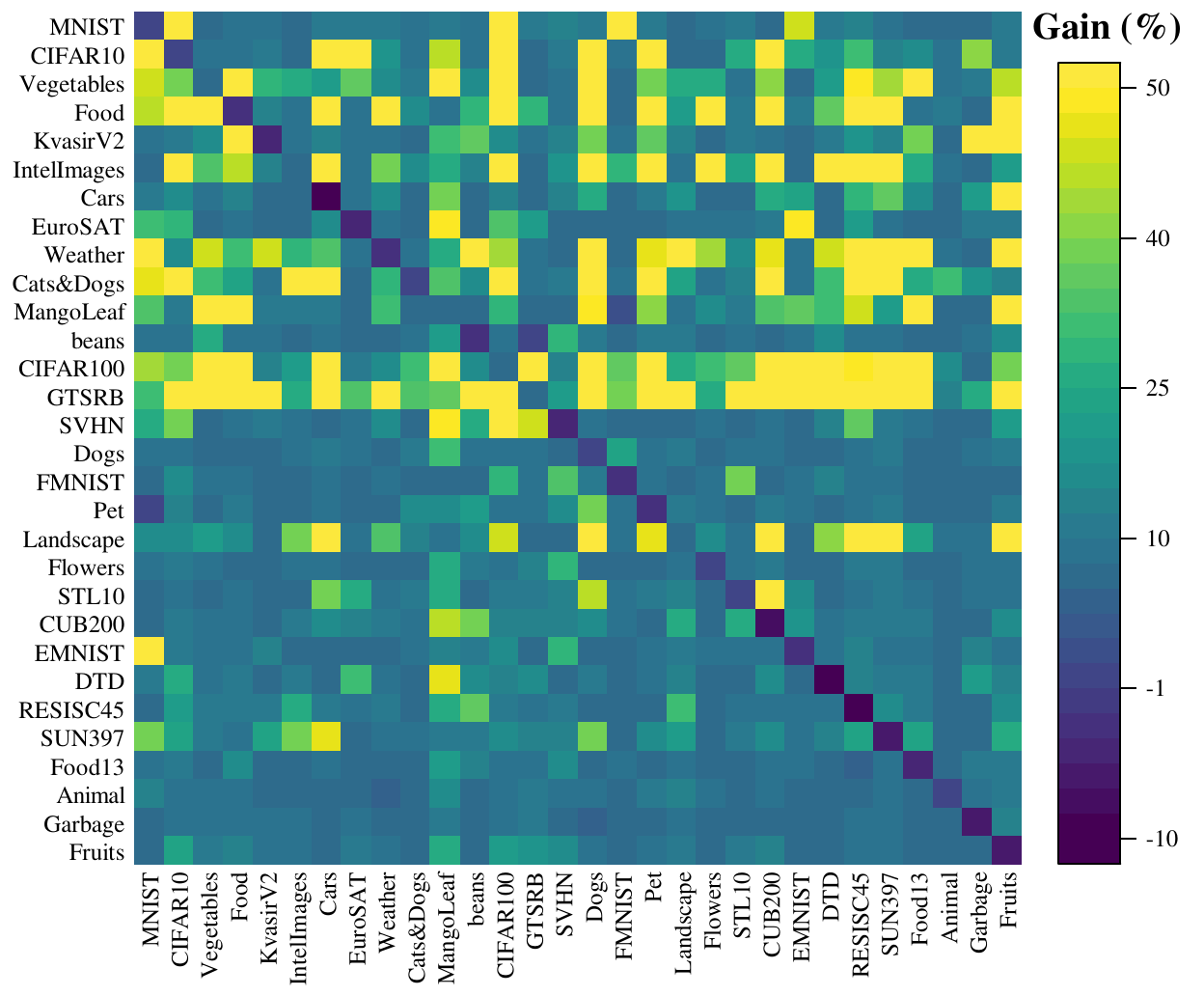} 
\vspace{-0.2cm}
\caption{High-pass filtering improves performance over the unfiltered version. Rows represent the added task vectors, while columns are the test datasets. The results show a significant boost in generalization with only a slight decline in task performance.} 
\label{figure:fourier.}  
\end{figure} 

Experimentally, we validate that high-pass filtering enhances model generalization with minimal performance loss as shown in Fig~\ref{figure:fourier.}. We fine-tune ViT-B/16~\cite{radford2021learning} on 30 visual tasks following~\cite{huang2024emr}. We find that removing low-frequency information results in a smaller performance drop (diagonal) compared to the improvement in generalization (off-diagonal). This indicates that low-frequency parts carry task-specific information that boosts task performance but reduces generalization. Thus, applying filtering like FR-Merging helps alleviate task interference during merging.

To maintain consistent output after merging, we define the merging coefficient $\lambda_i$ for the task vector $v_i$ as:
\begin{align}\label{eq:lambda}
    \lambda_i = \mathbb{E}(v_i)(\sum_{j=1}^K \mathbb{E}(v_j))^{-1},
\end{align}
where $\mathbb{E}$ represents the average. Overall, we construct a merged backbone in this part that minimizes task conflicts.

\subsection{Lightweight Expert Extraction}\label{sec:expert}
To compensate for information lost due to discarding low-frequency signals, we utilize lightweight experts added during inference. As analyzed in Sec.~\ref{sec:fr-merging}, low-frequency signals contain task-specific information, but preserving them requires an inefficient inverse Fourier transform during each inference. Thus, we propose a more efficient way.

Since fine-tuning leads to task specialization, the most changed parameters can be approximated as having the greatest impact~\cite{ma2023llm}. After reasonable rescaling, these selected parameters serve as task experts, minimizing storage (typically around 1\% of parameters) without extra inference computation. The validity is detailed in Appendix~\ref{appendix:dis:task experts}.

Although magnitude-based extraction~\cite{yadav2024ties} exists, limited research on rescaling hinders compression. As rescaling is crucial to expert extraction, we propose a rescaling method. Let $M(v_i,d)$ represent the top-$d$ ($d$ is a percentage) parameters for task vector $v_i$,  $\mu_i$ be the rescale factor, and $e(v_i)$ be the expert for task $t_i$. Then we have:
\begin{align}\label{equa: expert weight}
    e(v_i)=\mu_i M(v_i,d),\ \mu_i = -\frac{\mathbb{E}(M(v_i, d)) \cdot \log(d)}{\lambda_i \cdot \mathbb{E}(v_i) }.
\end{align}

The goal is to maintain a consistent output after extraction to preserve performance. Details are in Appendix~\ref{appendix:dis:Approximation_task experts}.

In summary, we efficiently extract task-specific experts to integrate into the backbone, mitigating task conflicts without excessive storage or inference overhead.

\begin{table*}[!ht]
    \centering
    \caption{Comparison of model merging performance using ViT-B/32 (B) and ViT-L/14 (L) on eight visual tasks evaluating by accuracy.}\label{tab:ViT-8-merging}
    \vspace{-0.15cm}
    \scalebox{0.78}{
    \begin{tabular}{l|c|cc|cc|cc|cc|cc|cc|cc|cc|cc}
    \toprule
        \multirow{2}{*}{Method} & \multirow{2}{*}{\makecell{Base\\ Model}} & \multicolumn{2}{c|}{SUN397} & \multicolumn{2}{c|}{Cars} & \multicolumn{2}{c|}{RESISC45} & \multicolumn{2}{c|}{EuroSAT} & \multicolumn{2}{c|}{SVHN} & \multicolumn{2}{c|}{GTSRB} & \multicolumn{2}{c|}{MNIST} & \multicolumn{2}{c|}{DTD} & \multicolumn{2}{c}{Avg.}\\
        \cline{3-20}
         & & B & L & B & L & B & L & B & L & B & L & B & L & B & L & B & L & B & L \\ \midrule
        \multicolumn{2}{l|}{Individual} & 75.3 & 82.3 & 77.7 & 92.4 & 96.1 & 97.4 & 99.7 & 100 & 97.5 & 98.1 & 98.7 & 99.2 & 99.7 & 99.7 & 79.4 & 84.1 & 90.5 & 94.2 \\ 
        \multicolumn{2}{l|}{Traditional MTL} & 73.9 & 80.8 & 74.4 & 90.6 & 93.9 & 96.3 & 98.2 & 96.3 & 95.8 & 97.6 & 98.9 & 99.1 & 99.5 & 99.6 & 77.9 & 84.4 & 88.9 & 93.5 \\ \midrule
        \multicolumn{2}{l|}{Weight Averaging} & 65.3 & 72.1 & 63.4 & 81.6 & 71.4 & 82.6 & 71.7 & 91.9 & 64.2 & 78.2 & 52.8 & 70.7 & 87.5 & 97.1 & 50.1 & 62.8 & 65.8 & 79.6 \\
        \multicolumn{2}{l|}{Fisher Merging~\cite{matena2022merging}} & \textbf{68.6} & 69.2 & \textbf{69.2} & \textbf{88.6} & 70.7 & 87.5 & 66.4 & 93.5 & 72.9 & 80.6 & 51.1 & 74.8 & 87.9 & 93.3 & 59.9 & 70.0 & 68.3 & 82.2 \\ 
        \multicolumn{2}{l|}{Task Arithmetic~\cite{ilharco2022editing}} & 63.8 & 74.1 & 62.1 & 82.1 & 72.0 & 86.7 & 77.6 & 93.8 & 74.4 & 87.9 & 65.1 & 86.8 & 94.0 & 98.9 & 52.2 & 65.6 & 70.1 & 84.5 \\ 
        \multicolumn{2}{l|}{Ties-Merging~\cite{yadav2024ties}} & 64.8 & 76.5 & 62.9 & 85.0 & 74.3 & 89.3 & 78.9 & 95.7 & 83.1 & 90.3 & 71.4 & 83.3 & 97.6 & 99.0 & 56.2 & 68.8 & 73.6 & 86.0 \\  \multicolumn{2}{l|}{Breadcrumbs~\cite{davari2023model}} & 63.0 & 74.9 & 61.0 & 84.9 & 75.7 & 88.6 & 84.7 & 95.4 & 83.4 & 89.8 & 76.4 & 90.1 & 98.1 & 99.2 & 57.9 & 68.2 & 75.0 & 86.4 \\ 
        \multicolumn{2}{l|}{PCB-Merging~\cite{du2024parameter}} & 65.5 & 75.8 & 64.1 & 86.0 & 78.1 & 88.6 & 80.2 & 96.0 & 84.7 & 88.0 & 77.1 & 90.9 & 98.0 & 99.1 & 58.4 & 70.0 & 75.8 & 86.9 \\ 
        \multicolumn{2}{l|}{FR-Merging(ours)} & 66.2 & \textbf{76.4} & 64.5 & 87.0 & \textbf{77.2} & \textbf{90.2} & \textbf{90.1} & \textbf{96.8} & \textbf{85.4} & \textbf{92.0} & \textbf{82.3} & \textbf{92.8} & \textbf{98.5} & \textbf{99.3} & \textbf{60.0} & \textbf{71.5} & \textbf{78.1} & \textbf{88.3} \\ \midrule
        
        \multicolumn{2}{l|}{AdaMerging++~\cite{yangadamerging}} & 66.6 & 79.4 & 68.3 & 90.3 & 82.2 & 91.6 & 94.2 & 97.4 & 89.6 & 93.4 & 89.0 & 97.5 & 98.3 & 99.0 & 60.6 & 79.2 & 81.1 & 91.0 \\ 
        \multicolumn{2}{l|}{Twin-Merging~\cite{lu2024twin}} & 73.6 & 82.6 & 71.7 & 90.3 & 92.1 & 94.9 & 99.3 & 99.4 & 95.3 & 96.7 & 97.2 & 98.0 & 99.1 & 99.4 & 74.0 & 80.1 & 87.8 & 92.7 \\ 
        \multicolumn{2}{l|}{EMR-Merging~\cite{huang2024emr}} & 74.1 & 82.3 & 72.7 & 90.8 & 91.9 & 95.3 & 99.4 & 99.5 & 95.8 & 96.9 & 96.9 & 98.1 & 99.1 & 99.5 & 72.1 & 80.1 & 87.7 & 92.8 \\ 
        \multicolumn{2}{l|}{FREE-Merging(ours)} & \textbf{77.1} & \textbf{83.5} & \textbf{78.2} & \textbf{92.4} & \textbf{93.4} & \textbf{96.2} & \textbf{99.5} & \textbf{99.6} & \textbf{96.3} & \textbf{98.1} & \textbf{98.2} & \textbf{98.7} & \textbf{99.5} & \textbf{99.7} & \textbf{75.4} & \textbf{81.7} & \textbf{89.7} & \textbf{93.7} \\ \bottomrule
    \end{tabular}
    }
    \vspace{-0.05cm}
    
\end{table*}

\subsection{FREE-Merging}\label{sec:free_merging}
\begin{algorithm}[tb]
    \caption{Workflow of FREE-Merging}
    \label{alg:algorithm1}
    \begin{algorithmic}[1] 
        \REQUIRE Task vectors $\{v_i\}_{i=1}^K$,
        pre-trained model $\theta_{pre}$, router $R$, and input data $X$.
        \STATE \textbf{FR-Merging:} \COMMENT{{\scriptsize Only excute once.}}
        \STATE Get backbone $\theta_m = \theta_{pre}+\sum_{i=1}^K \lambda_i G(v_i) $, where   $G$ and $\lambda$ are defined by Eq.~\ref{eq: fourier} and Eq.~\ref{eq:lambda}, respectively.
        \STATE \textbf{Expert Extraction:}  \COMMENT{{\scriptsize Only excute once.}}
        \STATE Get task experts $\{e_i\}_{i=1}^K$ as Eq.~\ref{equa: expert weight}.
        \STATE \textbf{Inference:}
        \FOR{$x \in X$}
                \STATE $[w_1,...w_K] \leftarrow argmax(R(x))$
                \STATE $\theta_*=\theta_m+\sum_{i=1}^Kw_i e_i$
                \STATE $Y \leftarrow Y \cup f(x;\theta_*)$
        \ENDFOR
        \ENSURE  Output $Y$ for input $X$.
    \end{algorithmic}
\end{algorithm}
This section presents FREE-Merging, combining both parts to make it suitable for multiple tasks without affecting inference speed. Inspired by MoE~\cite{shazeer2016outrageously}, a router dynamically assigns tasks to activated experts. The router can take various forms~\cite{lepikhingshard}, allowing for more flexibility in handling different tasks. Detailed discussions about router can be found in Appendix~\ref{appendix:dis:Role_of_Router}. For detailed algorithms, see Alg~\ref{alg:algorithm1}.

First, we obtain the merged backbone (line 2) and the task experts (line 4). Then, during the inference, we dynamically route the experts based on inputs (lines 7-9).

\subsection{Time Complexity and Storage Space Analysis}\label{sec:method:analysis}
This section analyzes the merging time complexity and storage requirements. We aim to minimize both for practical use. Assume there are $n$ models to merge, each with $m$ parameters and requiring storage space $s$.

The time complexity includes the merging and inference part. In the merging part, FR-Merging has a complexity of $O(nmlogm)$~\cite{duhamel1990fast} while expert extraction is $O(nm)$, making the overall merging complexity $O(nmlogm)$, which is manageable. For instance, merging three 10B models requires only $O(10^{12})$ addition operations, a minor load for modern devices. Moreover, with only a lightweight router added, the extra inference cost is negligible.

Next, we analyze storage cost. We typically save only about 1\% additional parameters per model, requiring $(1+0.01n)s$ storage, which is far less than the requirement of $ns$ for the model ensemble. For large models, reducing storage requirements is crucial for alleviating storage pressure on edge users. For detailed analysis, please refer to Sec~\ref{sec:speed and space}.

\section{Experiments}\label{sec:experiments}
In this section, we evaluate the proposed methods across various tasks. We compare FR-Merging with other cost-free methods (which do not require additional training or data), including Weight Averaging, Fisher Merging~\cite{matena2022merging}, Task Arithmetic~\cite{ilharco2022editing}, Ties-Merging~\cite{yadav2024ties}, Breadcrumbs~\cite{davari2023model}, and PCB-Merging~\cite{du2024parameter}. We also compare our FREE-Merging with popular methods, including RegMean~\cite{jin2022dataless}, AdaMerging~\cite{yangadamerging}, EMR-Merging~\cite{huang2024emr}, and Twin-Merging~\cite{lu2024twin}. The original EMR-Merging uses a perfect router; however, we use an imperfect router for realism. Some experiments include traditional MTL with data from all tasks as a baseline. The evaluation covers CV, NLP, and multi-modal tasks. 


\subsection{Merging Vision Models}\label{sec:exp:vision}

\noindent\textbf{Merging 8 ViTs.} We select the two visual encoders of CLIP, ViT-B/32 and ViT-L/14~\cite{radford2021learning}, as the pre-trained models and choose eight image classification tasks as target tasks: SUN397, Cars, RESISC45, EuroSAT, SVHN, GTSRB, MNIST, and DTD~\cite{yangadamerging}. Additionally, an MLP is used as the lightweight router. Detailed information on the datasets and baselines can be found in the Appendix~\ref{appendix:datasets}.

\begin{table}[!ht]
    \centering
     \caption{Comparison of model merging results using ViT-B/16 on 30 visual tasks. \textit{Additional Cost} indicates whether additional training or additional data is required during model merging.}\label{tab:ViT-30-avg}
     \vspace{-5pt}
    \scalebox{0.9}{
    \begin{tabular}{lcc}
    \hline
        Method &Additional Cost& Avg. Acc \\ \hline
        Individual& \ding{55} & 93.02 \\ \hline
        Weight Averaging& \ding{55} & 42.51 \\ 
        Task Arithmetic~\cite{ilharco2022editing}& \ding{55} & 48.88 \\ 
        Ties-Merging~\cite{yadav2024ties}& \ding{55} & 37.53 \\ 
        Breadcrumbs~\cite{davari2023model}& \ding{55} & 43.57 \\ 
        FR-Merging(ours)& \ding{55} & \textbf{53.90} \\ \hline
        RegMean~\cite{jin2022dataless}&\checkmark & 68.13 \\ 
        AdaMerging~\cite{yangadamerging}&\checkmark & 60.24 \\ 
        Twin-Merging~\cite{lu2024twin}&\checkmark & 75.52 \\ 
        EMR-Merging~\cite{huang2024emr}&\checkmark & 76.39 \\ 
        FREE-Merging(ours)&\checkmark & \textbf{79.67} \\ \hline
    \end{tabular}
    }
   \vspace{-5pt}
\end{table}

\begin{table}[t]
    \centering
    \caption{Language model merging results: RoBERTa fine-tuned on 8 discriminative tasks, T0-3B on 11 discriminative tasks with $IA^3$, and Qwen-14B on 3 generative tasks with LoRA.}\label{tab:lm_avg}
    \vspace{-5pt}
    \scalebox{0.85}{
    \begin{tabular}{l|c|c|cc|c}
    \toprule
        \multirow{2}{*}{Method}&\multirow{2}{*}{\makecell{Base \\ Model}}& \textbf{Full} & \multicolumn{2}{c|}{\textbf{PEFT}} & \multirow{2}{*}{\textbf{Avg.}}\\ 
         & & RoBERTa & T0-3B & Qwen-14B &\\ \midrule
        \multicolumn{2}{l|}{Individual} & 85.55 & 71.35 & 72.07 &76.32 \\ \midrule
        \multicolumn{2}{l|}{Weight Averaging} & 51.34 & 58.01 & 66.80 & 58.71 \\ 
   \multicolumn{2}{l|}{ Task Arithmetic\cite{ilharco2022editing}} & 66.65 & 63.91 & 66.40 &65.65 \\ 
        \multicolumn{2}{l|}{Ties-Merging~\cite{yadav2024ties}} & 64.04 & 66.41 & 67.46 &65.97\\ 
        \multicolumn{2}{l|}{Breadcrumbs~\cite{davari2023model}} & 68.19 & 54.64 & 66.77 &63.20\\ 
        \multicolumn{2}{l|}{PCB-Merging~\cite{du2024parameter}} & 67.86 & 66.10 & 66.69&66.39 \\ 
        \multicolumn{2}{l|}{FR-Merging(ours)} & \textbf{70.02} & \textbf{66.88} & \textbf{68.00} &\textbf{68.30}\\ \midrule
        \multicolumn{2}{l|}{RegMean~\cite{jin2022dataless}} & 70.01 & 58.01 & 67.52 &65.18\\ 
        \multicolumn{2}{l|}{EMR-Merging~\cite{huang2024emr}} & 74.20 & 67.11 & 70.98 &70.76\\ 
        \multicolumn{2}{l|}{Twin-Merging~\cite{lu2024twin}} & 78.29 & 66.70 & 71.68 &72.22\\ 
        \multicolumn{2}{l|}{FREE-Merging(ours)} & \textbf{80.16} & \textbf{68.68} & \textbf{72.78} &\textbf{73.87}\\ \bottomrule
    \end{tabular}
    }
    \vspace{-8pt}
\end{table}
Table~\ref{tab:ViT-8-merging} presents the comparative results. First, FR-Merging outperforms other cost-free methods, improving ViT-B/32 by 2.5\% and ViT-L/14 by 1.5\% over PCB-Merging~\cite{du2024parameter}, demonstrating its effectiveness in reducing performance loss caused by task interference. Through FR-Merging, we can construct a merged backbone that can be used independently at a low cost.  Here, we focus on the basic FR-Merging. In Sec.~\ref{sec:Transferability Analysis}, we discuss further improvements achieved by combining it with existing methods. Then, compared to methods requiring extra information, FREE-Merging also improves performance by 2\%, leveraging lightweight computation and efficient expert extraction, requiring only 1\% additional storage per model, outperforming 3\% of EMR-Merging~\cite{huang2024emr} and 2\% of Twin-Merging~\cite{lu2024twin}. Thus, FREE-Merging provides an excellent balance between storage, computation, and performance.

\noindent \textbf{Merging 30 ViTs.} We select ViT-B/16 as the pre-trained model and evaluate the model merging performance across 30 image classification tasks following~\cite{huang2024emr}. 

Table~\ref{tab:ViT-30-avg} presents the results, with detailed analysis provided in Appendix~\ref{appendix:additional-30ViTs}. It is evident that our proposed FR-Merging and FREE-Merging exhibit remarkable performance.  Although merging a large number of models is uncommon in practice, this experiment underscores the effectiveness of Fourier filtering and lightweight task experts in mitigating performance loss caused by task interference. These results highlight the robustness and broad applicability of our methods across diverse scenarios.

\subsection{Merging Language Models}\label{sec:exp:language}

\noindent\textbf{Merging Medium-sized Language Models.} We investigate the generalization ability of our method on medium-scale language models, using RoBERTa~\cite{zhuang-etal-2021-robustly} as the pre-trained model with full fine-tuning on eight discriminative tasks~\cite{yadav2024ties}. Details can be found in Appendix~\ref{appendix:datasets}.

The average merging results of the eight tasks are presented in Table~\ref{tab:lm_avg}, with complete results in Appendix~\ref{appendix:merging_language_models}. It can be observed that,  FR-Merging outperforms other cost-free methods by nearly 2\%. Additionally, FREE-Merging also exhibits remarkable effectiveness, achieving optimal results on almost all datasets and surpassing current methods in average performance. This highlights the great potential of our proposed methods for language models.

\noindent\textbf{Merging PEFT models.} PEFT is a popular method that reduces fine-tuning costs~\cite{ding2023parameter}, but storing PEFT parameters for multiple tasks remains space-inefficient~\cite{zhang2023composing}. The effectiveness of our approach on PEFT demonstrates the generalization capability and offers an efficient deployment way.

Firstly, we use T0-3B~\cite{sanhmultitask} as the pre-trained model and $(IA)^3$~\cite{liu2022few} as the PEFT method, fine-tuning on 11 discriminative tasks~\cite{huang2024emr}. Our proposed FR-Merging and FREE-Merging improve performance by nearly 1\%, as shown in Table~\ref{tab:lm_avg}. Next, we employ Qwen-14B~\cite{bai2023qwen} with LoRA~\cite{hulora} as the PEFT way, fine-tuning on three generative tasks~\cite{lu2024twin}. The results show that our methods lead to a 2\% advantage in both cost-free and full frameworks. This confirms the feasibility with PEFT and strong generalization of our methods. Full results are available in the Appendix~\ref{appendix:merging_language_models}.

\begin{table}[t]
    \centering
    \caption{Comparison of performance of different model merging methods after full fine-tuning LLaMa2-13B on three tasks: Instruction Following~(LM), Math, and Code Generation.}\label{tab:llm-merging}
    \scalebox{0.83}{
    \begin{tabular}{lcccc}
    \hline
        Method & AlpacaEval & GSM8K & MBPP& Avg. \\ \hline
        LM Fine-tuned~\cite{xu2024wizardlm} & 88.9 & 45.9 & 31.4&55.4 \\ 
        Math Fine-tuned~\cite{luo2023wizardmath} & 22.3 & 63.5 & 23.0&36.5 \\ 
        Code Fine-tuned~\cite{luowizardcoder} & 16.2 & 18.8 & 27.0&20.7 \\ \hline
        Task Arithmetic~\cite{ilharco2022editing} & 72.1 & 48.3 & 0&40.1 \\ 
        TIES-Merging~\cite{yadav2024ties} & 77.5 & 66.9 & 27.2&57.2 \\ 
        DARE~\cite{yu2024language} & 77.5 & 62.7 & 29.4&56.5 \\ 
        DELLA-Merging~\cite{deep2024della} & 80.4 & 61.8 & 31.4&57.9 \\ 
        PCB-Merging~\cite{du2024parameter} & 81.2 & 60.6 & 30.7&57.5 \\ 
        FR-Merging(ours) & \textbf{87.5} & \textbf{64.3} & \textbf{32.8}&\textbf{61.5} \\ \hline
        Twin-Merging~\cite{lu2024twin} & 83.5 & 61.4 & 30.4&58.4 \\ 
        EMR-Merging~\cite{huang2024emr} & 82.6 & 62.3 & 31.2&58.7 \\ 
        FREE-Merging(ours) & \textbf{88.1} & \textbf{65.7} & \textbf{31.8}&\textbf{61.8} \\ \hline
    \end{tabular}
    }
    \vspace{-0.1cm}
    
\end{table}
\begin{table*}[!ht]
\centering
\caption{Results of merging multi-modal BEiT3 models on five vision-language tasks.}\label{tab:performance_beit}
\scalebox{0.73}{
\begin{tabular}{lc|c|cccc|c|c|c}
\toprule
\multirow{2}{*}{Methods} &\multicolumn{1}{|c|}{Task} &\textbf{COCO-Retrieval} & \multicolumn{4}{c|}{\textbf{COCO-Captioning}} & \textbf{ImageNet-1k Classification} & \textbf{NLVR2}& \textbf{VQAv2}
\\
& \multicolumn{1}{|c|}{Metric}& Accuracy($\uparrow$) & BLEU4($\uparrow$) &CIDEr($\uparrow$)&METEOR($\uparrow$) & ROUGE-L($\uparrow$)& Accuracy($\uparrow$) & Accuracy($\uparrow$)  & Accuracy($\uparrow$)
\\
\midrule
\multicolumn{2}{l|}{Individual} & 0.8456 & 0.394 &1.337&0.311&0.601& 0.8537 & 0.7765 &0.8439 
\\
\midrule 
\multicolumn{2}{l|}{Weight Averaging} & 0.1893 & 0.031	&0.001	&0.115	&0.159 &  0.6771 & 0.2800 &  0.6285
\\
\multicolumn{2}{l|}{Task Arithmetic~\cite{ilharco2022editing}}  & 0.3177 & \textbf{0.033}	&0.000&	\textbf{0.118}&	\textbf{0.176} & 0.6732 & 0.3809 & 0.6933
\\
\multicolumn{2}{l|}{Ties-Merging~\cite{yadav2024ties}}& 0.3929 & 0.029&	0.001&	0.108	&0.167 & 0.6978 & 0.3206& 0.6717
\\
\multicolumn{2}{l|}{Breadcrumbs~\cite{davari2023model}}& 0.1893 & 0.011&	0.012&	0.065	&0.114 & 0.6323 & 0.6073& 0.6823
\\
\multicolumn{2}{l|}{FR-Merging(ours)}& \textbf{0.7010} & 0.016&	\textbf{0.019}&	0.096	&0.141 & \textbf{0.6992} & \textbf{0.6407}& \textbf{0.6965}
\\
\midrule
\multicolumn{2}{l|}{EMR-Merging~\cite{huang2024emr}} & 0.7946 & 0.289& 	1.060&	0.272&	0.534& 0.7742 & 0.7475 & 0.7211\\
\multicolumn{2}{l|}{Twin-Merging~\cite{lu2024twin}} & 0.7937 & 0.325& 	1.078&	0.278&	0.558& 0.7730 & 0.7762 & 0.7232\\
\multicolumn{2}{l|}{FREE-Merging(ours)} & \textbf{0.8093 }& \textbf{0.347}& 	\textbf{1.195}&	\textbf{0.289}&	\textbf{0.572}& \textbf{0.7850} & \textbf{0.8168} & \textbf{0.7297}
\\
\bottomrule
\end{tabular}
}
 
\end{table*}

\noindent\textbf{Merging Large Language models.} We explore large model merging using LLaMa2-13B~\cite{touvron2023llama} as the pre-trained model, with WizardLM~\cite{xu2024wizardlm} for instruction following, WizardMath~\cite{luo2023wizardmath} for math, and WizardCoder-Python~\cite{luowizardcoder} for coding. Experiments are conducted on AlpacaEval (instruction following), GSM8K (math), and MBPP (code)~\cite{yu2024language}. 

Table~\ref{tab:llm-merging} shows that our proposed FR-Merging even surpasses existing methods that rely on additional information. This indicates that our findings in FR-Merging regarding low-frequency signals and task interference also hold when extended to large models. This is crucial for practical usage of large models. Additionally, our proposed FREE-Merging also demonstrates strong advantages, enabling efficient cross-domain deployment of large models.

\subsection{Merging Multi-Modal Models}\label{sec:exp:multi-modal}
We merge various fine-tuned BEiT3-base~\cite{wang2023image} models across five tasks: ImageNet-1k (Classification), VQAv2 (Visual Question Answering), NLVR2 (Reasoning), COCO Captioning (Image Captioning), and COCO Retrieval (Image-Text Retrieval). COCO Captioning is evaluated with BLEU4, CIDEr, METEOR, and ROUGE-L, while the others use accuracy~\cite{huang2024emr}. Details are in Appendix~\ref{appendix:datasets}.

Table~\ref{tab:performance_beit} shows that our methods gain great improvements in multi-modal models. Unlike vision or language model merging, which often involve similar tasks, such as classification, the task similarity here is relatively low. The strong performance highlights the generalization ability of our methods, making them widely applicable.

\begin{figure}
\centering  
\includegraphics[width=0.95 \linewidth]{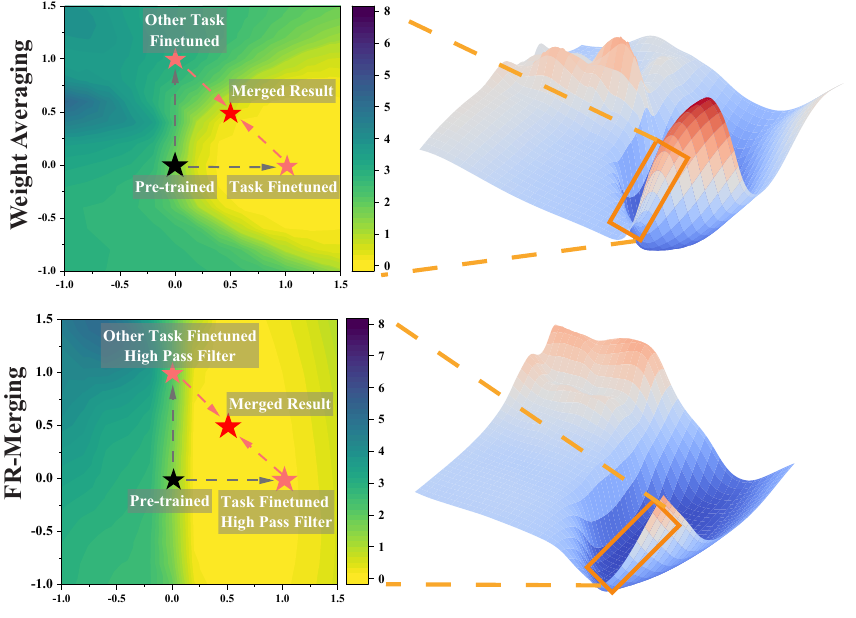}
\caption{Visualization of the loss landscape on EuroSAT before and after high-pass filtering~(top vs. bottom) in the model merging of the fine-tuned ViT-B/32 on EuroSAT and DTD tasks~\cite{ilharco2022editing}.} \label{figure:fourier effectiveness}  
\end{figure} 

\subsection{Fourier Transform Analysis}
In this section, we explain why Fourier high-pass filtering improves merging performance from an optimization perspective. Previous work~\cite{li2018visualizing} shows that landscape geometry impacts model generalization. Fig.~\ref{figure:fourier effectiveness} demonstrates that high-pass filtering widens the valley in the loss landscape (bottom plot). Since fine-tuned models from the same pre-trained model are closer in the landscape~\cite{neyshabur2020being}, a wider valley increases the probability of merged results falling within the loss basin as shown in the bottom plot of Fig.~\ref{figure:fourier effectiveness}, leading to better performance. Without filtering, the landscape is flat at a high level, which has little effect on model merging and only highlights the advantage of pre-training~\cite{garipov2018loss}.


\begin{table}[t]
    \centering
    \vspace{-0.2cm}
    \caption{Ablation of FREE-Merging components on 8 visual tasks.}\label{tab:ablation-all}
    \scalebox{0.65}{
    \begin{tabular}{ccccccc}
    \hline
        FR-Merging & Equa.3 & Top-K Expert & Scaling & ViT-B/32 & ViT-B/16 & ViT-L/14 \\ \hline
        \ding{55} & \ding{55} & \ding{55} & \ding{55} & 70.10 & 75.24 & 84.50 \\ 
        \ding{52} & \ding{55} & \ding{55} & \ding{55} & 76.96 & 81.34 & 88.21 \\ 
        \ding{55} & \ding{52} & \ding{55} & \ding{55} & 71.43 & 77.41 & 84.20 \\ 
        \ding{52} & \ding{52} & \ding{55} & \ding{55} & 78.10 & 82.54 & 88.30 \\ \hline
        \ding{55} & \ding{55} & \ding{52} & \ding{55} & 65.24 & 68.98 & 70.13 \\ 
        \ding{52} & \ding{52} & \ding{52} & \ding{55} & 79.38 & 84.31 & 89.53 \\ 
        \ding{55} & \ding{55} & \ding{52} & \ding{52} & 82.32 & 86.87 & 90.47 \\
        \ding{52} & \ding{52} & \ding{52} & \ding{52} & \textbf{89.68} & \textbf{91.09} & \textbf{93.70} \\ \hline
    \end{tabular}
    }
    \vspace{-0.05cm}
    
    \vspace{-0.1cm}
\end{table}
\subsection{Ablation Study}\label{sec:exp:ablation}
In this section, we conduct ablation experiments on the components of our method. All experiments are performed on ViT-B/32 with eight classification tasks as target tasks.

\begin{figure*}[htbp]
  \centering
  \begin{subfigure}{0.33\linewidth}
    \includegraphics[width=1\textwidth]{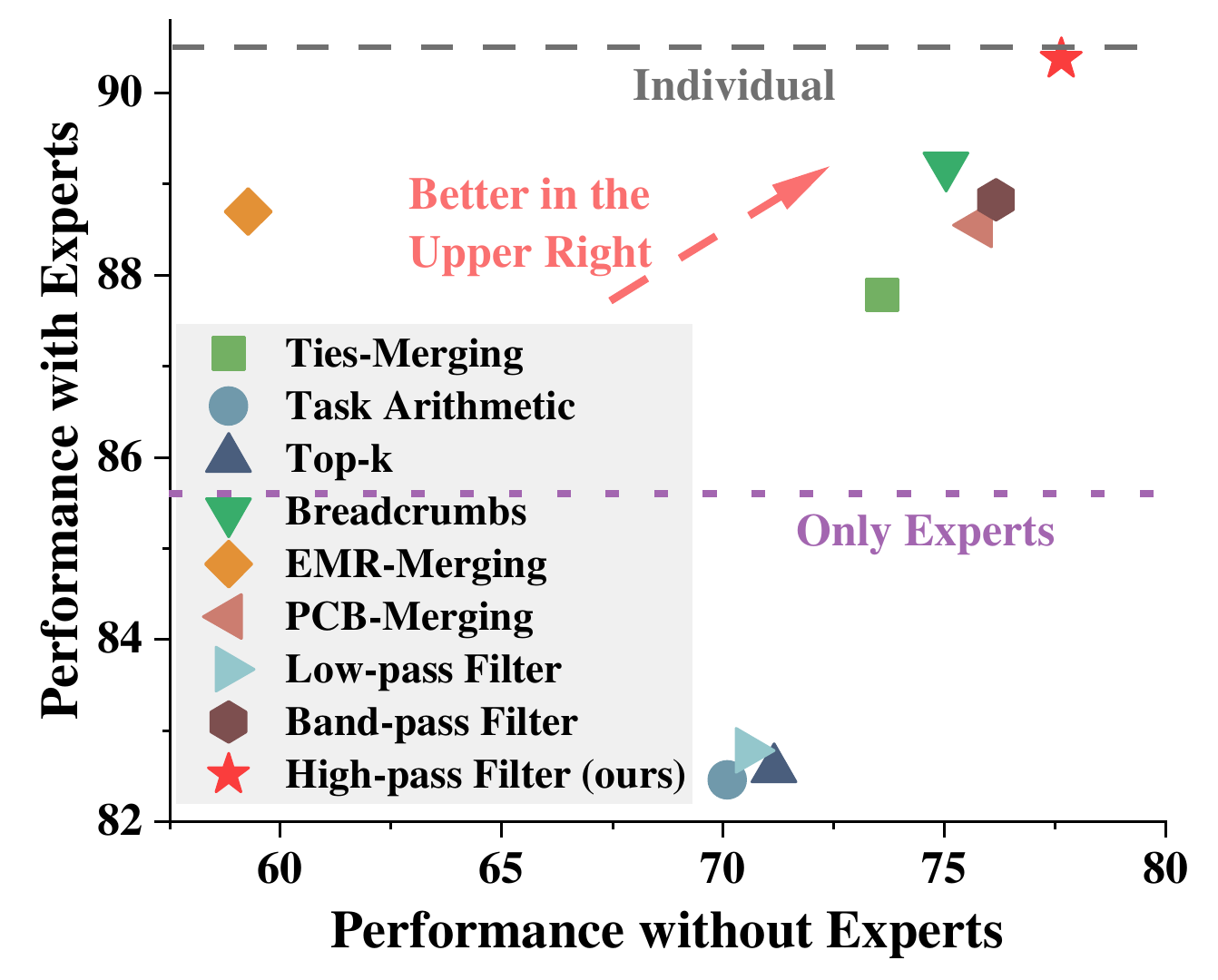}
    \caption{Ablation on FR-Merging.}
    \label{fig:ablation-fourier}
  \end{subfigure}
  \hfill
  \begin{subfigure}{0.33\linewidth}
    \includegraphics[width=1\textwidth]{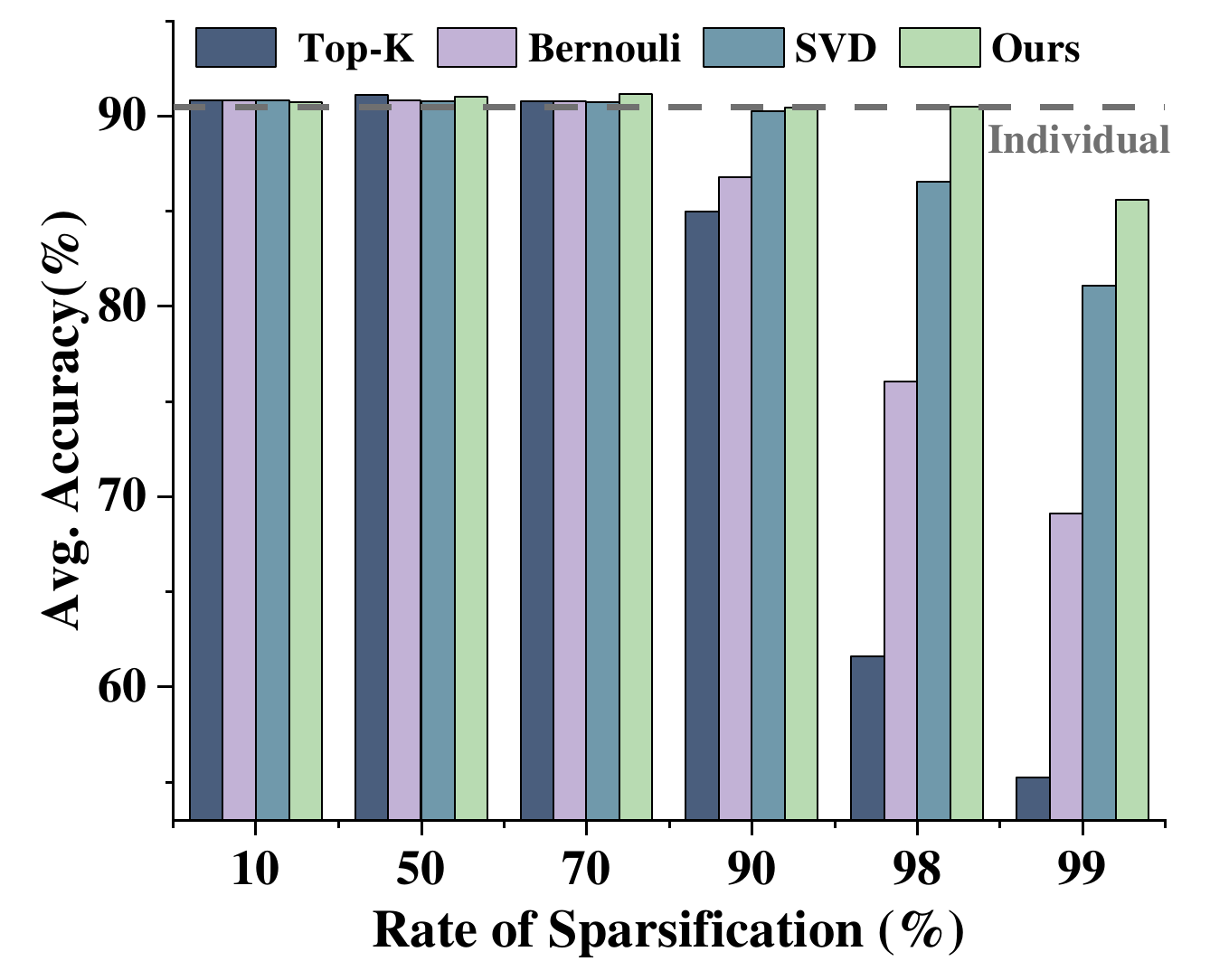}
    \caption{Ablation on Task Expert.}
    \label{fig:ablation-expert}
  \end{subfigure}
  \begin{subfigure}{0.33\linewidth}
    \includegraphics[width=1\textwidth]{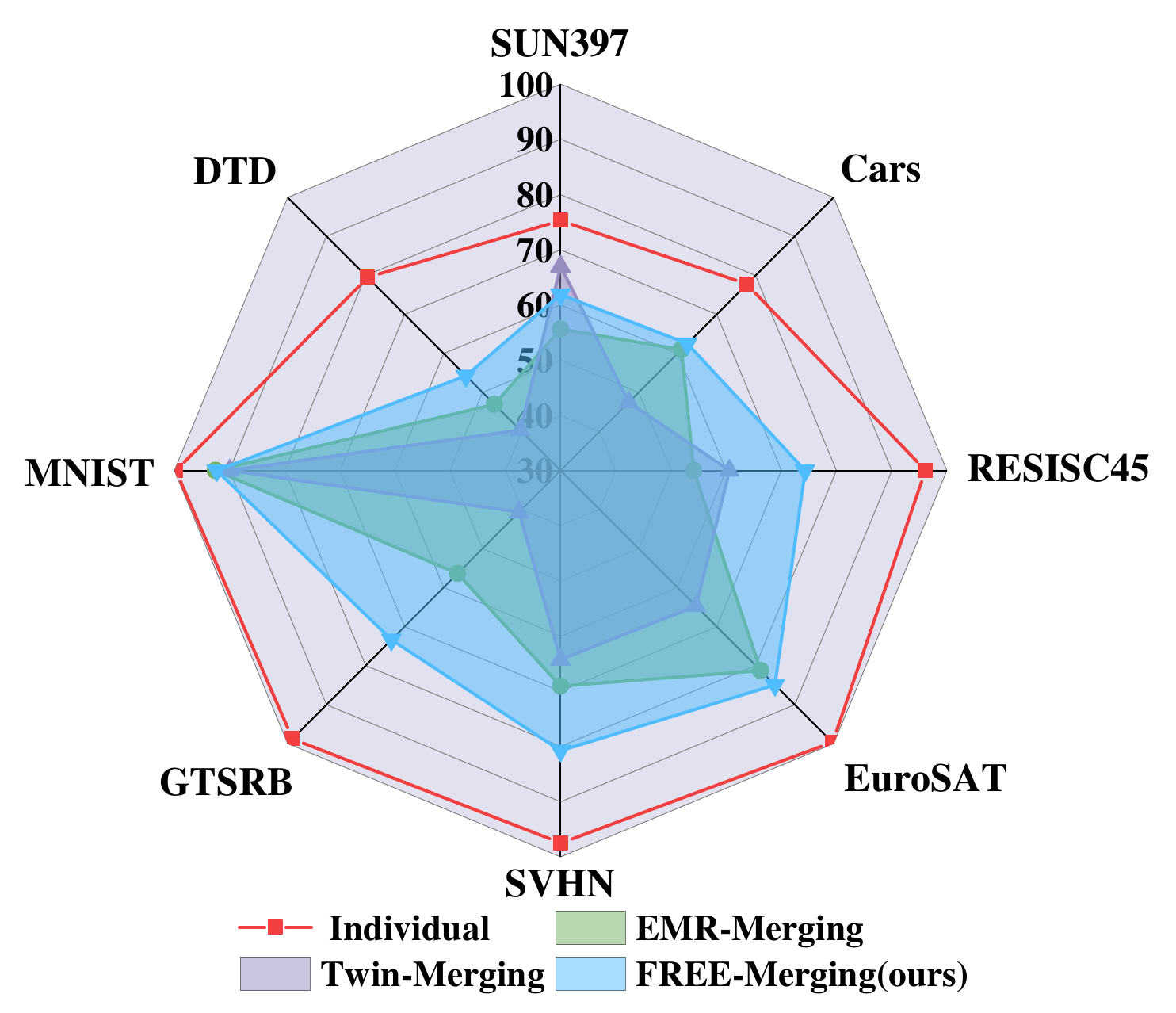}
    \caption{Router Analysis.}
    \label{fig:ablation-router}
  \end{subfigure}
  \vspace{-0.5cm}
  \caption{a) The ablation experiments on FR-Merging show that our proposed methods achieve optimal performance. b) Performance of experts at different sparsity rates. c) In experiments with random routers, our approach demonstrates optimal performance.}
  \label{fig:short}
  \vspace{-0.1cm}
\end{figure*}

\noindent\textbf{Ablation on FR-Merging.}
We analyze the impact of FR-Merging on the backbone and overall performance. Table~\ref{tab:ablation-all} shows that the performance of various models greatly improves after employing FR-Merging. We then compare high-pass filtering with other cost-free methods, including Ties~\cite{yadav2024ties}, Top-K, Task Arithmetic~\cite{ilharco2022editing}, outlier dropout~\cite{davari2023model}, EMR-Merging~\cite{huang2024emr}, and PCB-Merging~\cite{du2024parameter}, as well as low-pass and band-pass filtering, with results in Fig.~\ref{fig:ablation-fourier}. For fairness, all experts use 1\% additional parameters.


\begin{table}[t]
    \centering
    \caption{Transferability analysis of FR-Merging on 8 visual tasks.}\label{tab:transferability}
    \vspace{-0.2cm}
    \scalebox{0.9}{
    \begin{tabular}{cccc}
    \hline
        Method &FR-Merging & ViT-B/32 & ViT-L/14 \\ \hline
        Ties-Merging~\cite{yadav2024ties} & \ding{55} & 73.6 & 86.0 \\ 
        Ties-Merging~\cite{yadav2024ties} & \ding{52} & \textbf{78.3} & \textbf{88.4} \\ \hline
        Breadcrumbs~\cite{davari2023model}& \ding{55} & 75.0 & 86.4 \\ Breadcrumbs~\cite{davari2023model}& \ding{52} & \textbf{78.8} & \textbf{88.9}\\ \hline
        TSVM~\cite{gargiulo2024task}& \ding{55} & 83.9 & 91.9 \\ 
        TSVM~\cite{gargiulo2024task}& \ding{52} & \textbf{84.8} & \textbf{92.5}  \\ \hline
    \end{tabular}
    }
    \vspace{-0.3cm}
    
\end{table}

Our proposed FR-Merging outperforms other cost-free methods on both the backbone and with task experts, minimizing task interference. Examining the different filtering methods, we find that low-pass filtering leads to great performance degradation, confirming that task interference in low-frequency regions is severe. Band-pass filtering also performs worse than FR-Merging, and this is because some high-frequency signals preserve abrupt changes in the signal and their removal harms the performance of the model.

\noindent\textbf{Ablation on Merging Coefficients.} We conduct an ablation study on our merging coefficients as in Eq.~\ref{eq:lambda}, with results in Table~\ref{tab:ablation-all}. Using them instead of simple averaging yields slight improvements by considering output variations, benefiting tasks with varying fine-tuning magnitudes.

\noindent\textbf{Ablation on Task Expert.}
We conduct an ablation study on the expert extraction way. Table~\ref{tab:ablation-all} shows that incorporating our experts and rescaling method results in significant improvements, confirming their effectiveness. To validate the superiority of our method, we evaluate expert-only performance with varying parameter proportions, comparing Top-K selection, Bernoulli selection~\cite{yu2024language}, SVD extraction~\cite{lu2024twin}, and our Top-K selection with rescaling. Fig.~\ref{fig:ablation-expert} shows that our method excels, particularly under high sparsity. 

\noindent\textbf{Router Analysis.}
We analyze the proposed router method, aiming for strong merging performance despite sub-optimal routing. Since router inaccuracies may arise from limited training resources or edge deployment issues, our methods maintain performance in real-world scenarios.
Using a random router to simulate inaccuracy, we compare our method with existing router-based approaches. Our method outperforms Twin-Merging by 13.5\% and EMR-Merging by 8.5\% on average, as shown in Fig.~\ref{fig:ablation-router}, indicating that our methods construct a reasonable backbone and generalizable experts.
\subsection{Transferability Analysis}\label{sec:Transferability Analysis}

Our proposed FR-Merging can integrate with other methods, as it addresses frequency-domain task interference that current methods overlook. As shown in Table~\ref{tab:transferability}, combining various methods with FR-Merging leads to performance improvements. Therefore, FR-Merging can be integrated with more approaches to achieve greater benefits.
\subsection{Speed and Storage Space Analysis}\label{sec:speed and space}
\begin{table}[t]
    \centering
    \caption{Trade-off of performance, training, inference, and storage costs using ViT-L/14 across 8 image classification tasks.} \label{tab:speed storage}
    \vspace{-0.2cm}
    \scalebox{0.76}{
    \begin{tabular}{lccccc}
    \hline
        \multirow{2}{*}{Method} & \multirow{2}{*}{\makecell{Training \\ Tokens}} & \multirow{2}{*}{\makecell{Training \\Cost}} & \multirow{2}{*}{\makecell{Storage \\Param.}} & \multirow{2}{*}{\makecell{Inference Cost \\(/1000 items)}} & \multirow{2}{*}{Acc.} \\ 
        & & & & & \\ \hline
        Individual & 0 & 0 & 2.4B & 31.0s & 94.2 \\ \hline
        MTL & 304M & 12h & 304M & 31.0s & 93.5 \\ 
        FR-Merging  & 0 & 0 & 304M & 31.0s & 88.2 \\ 
        Twin-Merging & 0.9M & 147s & 352M & 32.2s & 92.7 \\ 
        FREE-Merging & 0.9M & 147s & 328M & 32.2s & 93.7 \\ \hline
    \end{tabular}
    }
    \vspace{-0.05cm}
    
    \vspace{-0.2cm}
\end{table}
This part analyzes training costs, inference speed, and storage as shown in Table~\ref{tab:speed storage}. FR-Merging achieves strong performance with significantly reducing storage costs without extra training or inference costs, while FREE-Merging slightly increases overhead for better performance. Both of these two methods enhance deep learning deployment.
\section{Conclusion}\label{sec:conclusion}
In this paper, we propose FR-Merging and FREE-Merging to efficiently build a multi-task model by merging existing ones. FR-Merging constructs a backbone without training by high-pass filtering to remove information that dominates task interference. Meanwhile, FREE-Merging employs lightweight experts to offset information loss during merging. Our method balances training, storage, inference costs, and performance, showing high practical value.
\section*{Acknowledgements}
This work is supported by National Natural Science Foundation of China (NSFC) (62232005, 62202126); the National Key Research and Development Program of China (2021YFB3300502). 

{
    \small
    \bibliographystyle{ieeenat_fullname}
    \bibliography{main}
}
\clearpage
\setcounter{page}{1}
\maketitlesupplementary

\appendix
\section{Notations}\label{appendix:notations}
We first list the notations for key concepts in our paper.
\begin{table}[!ht]
    \centering
\caption{Notations.}\label{tab:appendix:notations}
    \scalebox{0.8}{
    \begin{tabular}{ll}
    \hline
        Notations & Descriptions \\ \hline
        $f_\theta$ & A neural network with parameters $\theta$. \\
        $\theta_{pre}$ & Pre-trained model parameters. \\ 
        $\theta_k$ & Fine-tuning parameters for task $k$. \\ 
        $\theta_m$ & Model merging backbone network. \\ 
        $v_k$ & Task vector for task $k$, $v_k=\theta_k-\theta_{pre}$. \\ 
        \multirow{2}{*}{$M(v_k,d)$} & \multirow{2}{*}{\makecell{The top 100d\% parameters with the most \\significant changes corresponding to $v_k$.}} \\ 
        ~&~\\
        $\mu_k$ & The rescaling coefficient of the expert of task $k$\\
        $e_k$ & Lightweight task expert for task $k$, $e_k=\mu_k M(v_k,d)$. \\ 
        $\theta_*$ & The backbone with the corresponding expert. \\ 
        $G(\cdot)$ & Transformation operations on parameters. \\ 
        $\lambda_k$ & Merging coefficient of the model parameters for task $k$. \\ 
        
        $x \in R^m$ & The input of the neural network. \\ 
        $Y \in R^n$ & The output of the neural network. \\ 
        \hline
    \end{tabular} }
\end{table}
\section{Proof}\label{appendix:proofs}
In this section, we provide a proof for Theorem~\ref{theorem:moe need}. We begin by introducing the notation used in this proof.

\subsection{Notations and preliminaries}
To illustrate the proof, let's consider two fine-tuned models with parameters $\theta_A$ and $\theta_B$, For clarity, we focus on a single-layer MLP (without activation functions), though we will explain at the end how our proof generalizes to all cases. The merged result is given by: $\theta_m=\lambda_1G(\theta_A)+\lambda_2G(\theta_B)$, where $G(\cdot)$ represents a transformation operation applied to the parameters. Assume that $\theta_A$ corresponds to task input $X_A$, and $\theta_B$ corresponds to task input $X_B$.

The goal is to prove that, without additional storage or extra training for $\theta_A$ and $\theta_B$, simply storing $\theta_m$ cannot perfectly retain the capabilities of both $\theta_A$ and $\theta_B$.

\subsection{Details of proof}
Next, we will analyze the cases separately. 

\noindent\textbf{Without transformations.} Assume that we do not apply transformations to the output of the merged network. the output of the model $f_{\theta}$ can be expressed as:
\begin{equation}
    output_f=\theta x
\end{equation}

Suppose $\theta_m$ can perfectly retain the capabilities of both $\theta_A$ and $\theta_B$ without introducing new knowledge. That is, for $\forall x_A \in X_A, x_B \in X_B $, we have:
\begin{align}\label{appendix:prof:1}
\theta_m x_A= \theta_A x_A,\ \theta_m x_B= \theta_B x_B.
\end{align}

Then we have:
\begin{align}\label{appendix:prof:2}
[\lambda_1G(\theta_A)+\lambda_2G(\theta_B)]x_A=\theta_A x_A, \\
[\lambda_1G(\theta_A)+\lambda_2G(\theta_B)]x_B=\theta_B x_B.
\end{align}

Assume that $\exists \ x_m =\mu_1 x_A+\mu_2 x_B \in X_A$, where $\mu_1, \mu_2 \in \mathbb{N}$. Substituting, we can obtain:
\begin{align}
&\theta_m x_m=\theta_Ax_m\\
&\Rightarrow \theta_m (x_A-x_m)=\theta_A(x_A-x_m)\\
&\Rightarrow \theta_m[(1-\mu_1)x_A-\mu_2x_B]
\\& \quad=\theta_A[(1-\mu_1)x_A-\mu_2x_B]\\
&\Rightarrow \mu_2\theta_mx_B=\mu_2\theta_Ax_B\\
&\Rightarrow \mu_2\theta_Bx_B=\mu_2\theta_Ax_B \label{eq:14}
\end{align}

Due to:
\begin{equation}\label{eq:15}
    \mu_2\theta_mx_B=\mu_2\theta_Bx_B
\end{equation}
Thus, if the Eq.~\ref{eq:14} and Eq.~\ref{eq:15} holds for $\forall x_A \in X_A, x_B \in X_B $, then $\mu_2=0$. Therefore, we have:
\begin{equation}\label{eq:16}
    \forall \mu_2 \neq0, \nexists \ x_m=\mu_1 x_A+\mu_2 x_B \in X_A
\end{equation}
 
Eq.~\ref{eq:16} indicates that $X_A$ is not continuous. However, we usually consider the input space of the model to be continuous. Even if it is discontinuous, we expect the model to be robust to slight perturbations in the input. 

Thus, the assumption that $\theta_m$ can perfectly retain $\theta_A$ and $\theta_B$ without extra data or training does not hold in this case.

\noindent\textbf{With transformations.} Assuming we apply certain transformations to the output of the merged network, the output of the model $f_\theta$ can be expressed as:
\begin{equation}
    output_f=h(\theta x),
\end{equation}
where  $h(\cdot)$represents a transformation applied to the output.

By the universal approximation theorem, $h(\cdot)$ can be approximated by a two-layer MLP with sigmoid activation. Then, assuming $\theta_m$ can perfectly retain the capabilities of both $\theta_A$ and $\theta_B$. Then for $\forall x_A \in X_A, x_B \in X_B $, we have:
\begin{align}
W_2[sigmoid(W_1\theta_m x_A)]=\theta_A x_A\\
W_2[sigmoid(W_1\theta_m x_B)]=\theta_B x_B.
\end{align}
Substituting, we obtain:
\begin{align}
    \frac{W_2}{1+e^{-W_1\theta_mx_A}}=\theta_Ax_A,\\
\frac{W_2}{1+e^{-W_1\theta_mx_B}}=\theta_Bx_B.
\end{align}
Then we have:
\begin{align}\label{appendix:prof:W1}
W_2&=\theta_Ax_A(1+e^{-W_1\theta_mx_A})
    \\&=\theta_Bx_B(1+e^{-W_1\theta_mx_B}).
\end{align}
Thus, we can obtain the following equation:
\begin{align}
&\theta_Ax_A(e^{W_1\theta_mx_A}+1)e^{W_1\theta_mx_B}\\
&=\theta_Bx_B(e^{W_1\theta_mx_B}+1)e^{W1\theta_mx_A}.
\end{align}
Therefore,
\begin{align}\label{appendix:prof:W2}
    &(\theta_Ax_A-\theta_Ax_B)(e^{W_1\theta_mx_B}e^{W_1\theta_mx_A}+e^{W_1\theta_mx_B})
    \\&=\theta_B x_B(e^{W_1\theta_mx_B}+e^{W_1\theta_mx_A})
\end{align}
Therefore, based on Equa.~\ref{appendix:prof:W1} and Equa.~\ref{appendix:prof:W2}, since they hold for $\forall x_A \in X_A, x_B \in X_B $, at least one of  $W_1$ or $W_2$ can be expressed as $\Phi(\theta_A,\theta_B)$, where 
$\Phi$ represents a function.

In other words, only by additionally storing part of the information from $\theta_A$ or  $\theta_B$ and processing $\theta_m$ at the output can $\theta_m$ perfectly retain the capabilities of both $\theta_A$ and $\theta_B$. Therefore, the assumption that  $\theta_m$ can perfectly retain both $\theta_A$ and $\theta_B$ does not hold in this case.

Next, we briefly outline the rationale for extending this result to all layers of a neural network. Our assumptions involve only a single layer of the neural network, which already cannot perfectly retain capabilities without introducing new data. Therefore, it is even less feasible for the entire network to achieve this. Thus, this extension is reasonable. 

In summary, the proposition is proven.

\section{Reproducibility}\label{appendix:reproducibility}
\subsection{Datasets}\label{appendix:datasets}
\noindent \textbf{Merging 8 ViTs.} We use ViT-B/32 and ViT-L/14 as pre-trained models, and fine-tune them on 8 image classification datasets~(SUN397, Cars, RESISC45, EuroSAT, SHVN, GTSRB, MNIST, and DTD), then merge the models and test their performance. Configuration details follow~\cite{huang2024emr}.

\noindent \textbf{Merging 30 ViTs.} We use ViT-B/16 as the pre-trained model and test its merging performance on 30 image datasets. The datasets include MNIST, CIFAR-10, Vegetables, Food-101, Kvasir-v2, Cars, Intel Images, EuroSAT, Weather, Cats and Dogs, MangoLeafBD, Beans, CIFAR-100, GTSRB, SHVN, Dogs, Fashion MNIST, Oxford-IIIT-Pet, Landscape Recognition, Flowers Recognition, STL-10, CUB-200-2011, EMNIST, DTD, RESISC45, SUN397, KenyanFood13, Animal-10N, Garbage Classification, and Fruits-360. These datasets cover a wide range of major image categories. Configuration details follow~\cite{huang2024emr}.

\noindent \textbf{Merging Medium-sized Language Models.} We use RoBERTa as the pre-trained model, fine-tune it on 8 classification task datasets from GLUE benchmark for model merging, including CoLA, SST-2, MRPC, STS-B, QQP, MNLI, QNLI, and RTE. CoLA is evaluated with the Matthews correlation coefficient, STS-B with the average of the Pearson and Spearman correlation coefficients, and the others by accuracy. Details follow~\cite{huang2024emr}.

\noindent \textbf{Merging PEFT Models.} In this section, we conduct two experiments. The first uses T0-3B as the pre-trained model and $IA^3$ as the PEFT method, fine-tuned and merged on RTE, CB, Winogrande, WiC, WSC, COPA, H-SWAG, Story Cloze, and ANLI (R1 to R3). Details can be found in~\cite{huang2024emr}. The second part uses Qwen-14B as the pre-trained model and LoRA as the PEFT method. We fine-tune and merging on three generative tasks: MMLU, TruthfulQA, and BBQ. Configuration details can be found in~\cite{lu2024twin}.

\noindent \textbf{Merging Large Language Models.} We use LLaMa2-13B as the pre-trained model and apply fine-tuned results from WizardLM, WizardMath, and WizardCoder-Python for instruction following, math, and code tasks, respectively. We perform merging and test the model on three datasets: AlpacaEval (instruction following task), GSM8K (math task), and MBPP (code generation task). Details follow~\cite{yu2024language}.

\noindent \textbf{Merging Multi-Modal Models.} We use BEiT3 as the pre-trained model, fine-tune it on five multi-modal datasets, and test model merging performance. The datasets are ImageNet-1k (Image Classification), VQAv2 (Visual Question Answering), NLVR2 (Visual Reasoning), COCO Captioning (Image Captioning), and COCO Retrieval (Image-Text Retrieval). COCO Captioning is evaluated using BLEU4, CIDEr, METEOR, and ROUGE-L, while the other tasks are measured by accuracy. Details follow~\cite{huang2024emr}.

\subsection{Baselines}\label{appendix:baselines}
\noindent\textbf{Individual.} Solve each task with its fine-tuned model, but this requires storing separate models for each task, leading to significant storage overhead.

\noindent\textbf{Traditional MTL.} 
Train a model on data from all tasks, which incurs high training costs and data privacy issues.

\noindent\textbf{Weight Averaging.} Average the parameters for merging. The method is simple, but it faces great performance loss.

\noindent\textbf{Fisher Merging}~\cite{matena2022merging}. Use Fisher information matrices to assess parameter importance and determine merging coefficients. However, this requires calculating the matrix for each model, demanding high computational resources, making it unsuitable for edge deployment.

\noindent\textbf{Task Arithmetic}~\cite{ilharco2022editing}. Define task vectors as the merging target. For task $k$, the task vector is defined as $v_k=\theta_k-\theta_{pre}$, where $\theta_{pre}$ is the pre-trained model parameters, and $\theta_k$ is the fine-tuned parameters for task $k$. The merging process can be represented as $\theta_m=\theta_{pre}+\lambda \sum_{i=1}^K v_i$, where $\lambda$ is the merging coefficient. This method suffers significant performance degradation due to unaddressed task conflicts.

\noindent\textbf{Ties-Merging}~\cite{yadav2024ties}. Attempts to resolve parameter conflicts during model merging by eliminating redundancy and sign conflicts. However, resolving parameter conflicts is insufficient to address task conflicts, resulting in performance loss.

\noindent\textbf{Breadcrumbs}~\cite{davari2023model}. Discards parameters with the largest and smallest absolute values as redundant, negatively impacting model merging. This approach is simple, but it shows a significant performance decline on certain tasks.

\noindent\textbf{PCB-Merging}~\cite{du2024parameter}. Uses internal balancing to measure parameter importance within tasks and mutual balancing to assess parameter similarity across tasks, discarding redundant parameters and adjusting merging coefficients. 
This approach requires considerable computational resources, whereas our FR-Merging approach requires less computational power while yielding better results.

\noindent\textbf{RegMean}~\cite{jin2022dataless}. A weighted merging model based on a closed-form solution for the merging problem, aiming to maximize the similarity between the merged model and each pre-merged model on the same input. Requires inner product data from training. Since this method relies on data information, which is typically not provided by the model provider, its applicability is limited to specific scenarios.

\noindent\textbf{AdaMerging}~\cite{yangadamerging}. Uses an unsupervised approach to learn the merging coefficient for each task vector or layer. AdaMerging++ additionally applies Ties-Merging before calculating the merging coefficient. This method is limited to classification tasks and requires certain training resources, making it unsuitable for edge deployment.

\noindent\textbf{DARE}~\cite{yu2024language}. DARE randomly discards a large portion of task vector parameters before merging, potentially reducing parameter interference among models. This method is simple, but due to the lack of further optimization in the merging, it suffers from significant performance degradation.

\noindent\textbf{EMR-Merging}~\cite{huang2024emr}. Proposes retaining a mask matrix and rescale parameters for each model. During inference, the mask matrix and rescaling parameters are selected to recover performance. 
This method cannot further improve performance because it does not optimize the merged backbone. Additionally, since a mask must be stored for each task, it requires considerable storage space. In contrast, our method optimizes these aspects simultaneously.

\noindent\textbf{Twin-Merging}~\cite{lu2024twin}. Proposes saving an expert module extracted by SVD from the original parameters for each task, which is dynamically added during inference. This method does not optimize the merged backbone, resulting in suboptimal performance. It requires storing a large number of parameters when constructing task experts. In contrast, our proposed method optimizes both aspects, making it suitable for resource-constrained edge deployment scenarios.

\subsection{Details}\label{appendix:details}
In this section, we discuss the details of the experiments. All experiments, except for the speed tests, are conducted on eight NVIDIA RTX-3090 GPUs. The inference speed tests are performed using a single RTX-3090 GPU.

\begin{figure}[t] 
\centering  
\includegraphics[width=0.995 \linewidth]{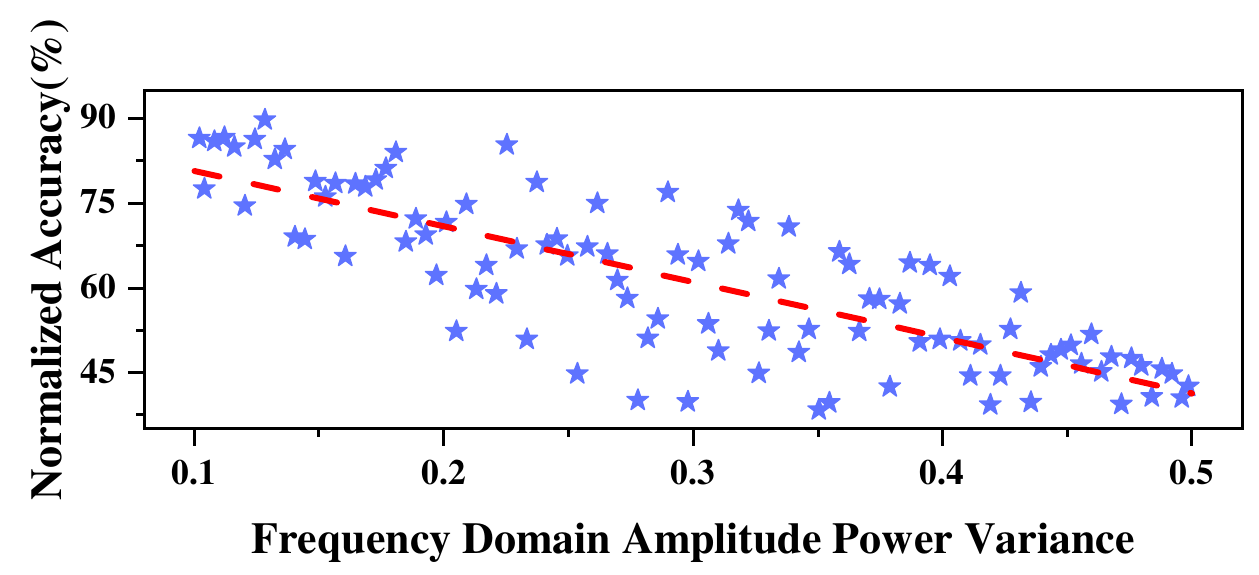} 
\vspace{-0.4cm}
\caption{The relationship between the merging performance of ViT-B-32 and the variance of frequency domain amplitude power. Normalized accuracy is the ratio of merged model performance to fine-tuned model performance on the same task.}\label{fig:t5_var_acc}
\vspace{-0.2cm}
\end{figure} 

\section{More Discussions and Future Directions}\label{appendix:discussion}
In this section, we first provide a more detailed discussion of task interference in the frequency domain in~\ref{appendix:task_inteference}. Next, we present the experimental results of FR-Merging on language models in~\ref{appendix:expanded_frequency_domain_analysis} and~\ref{appendix:fourier_transform_on_language}. Additionally, we discuss the effectiveness and construction of the expert module in~\ref{appendix:dis:task experts} and~\ref{appendix:dis:Approximation_task experts}, as well as the role of the router in~\ref{appendix:dis:Role_of_Router}. We also explore the selection of the cutoff frequency in~\ref{appendix:dis:cutoff_hyperparameter} and explored the implications of forgetting model-specific capabilities in~\ref{appendix:dis:Fuse_to_Forget}. Finally, we conclude with a discussion of future work in~\ref{appendix:future}.
\subsection{Task Interference in Frequency Domain}\label{appendix:task_inteference}

In this section, we further discuss the task interference presented in Sec.~\ref{sec:Task_Conflict} and provide visualization results.

The first point to address is why we consider the difference in low-frequency amplitude in the frequency domain as the cause of task interference. The amplitude in the frequency domain represents the strength of a signal. During the merging process, if one signal is strong while the other is weak, the weaker signal is likely to be overshadowed by the stronger one, regardless of their directional relationship. This manifests in the merged model as the near-complete loss of performance from one of the fine-tuned models. Therefore, a significant difference in the low-frequency amplitude will indeed lead to task interference during merging.

In Sec.~\ref{sec:Task_Conflict}, we quantify interference using the mean variance, observing a strong negative correlation with merged model performance on ViT-B/32. Here, normalized performance refers to the ratio of the performance of merged model on a task to that of the fine-tuned model on the same task. This negative correlation serves as strong evidence of task interference in the frequency domain.

Next, we verify whether this negative correlation also holds for language models. We choose Flan-T5~\cite{chung2024scaling} as the base model and conduct experiments on eight GLUE datasets. As shown in Fig.~\ref{fig:t5_var_acc}, the negative correlation is observed in language models, demonstrating that task interference in the frequency domain is a general phenomenon.

\begin{figure*}[t] 
\centering  
\includegraphics[width=0.995 \linewidth]{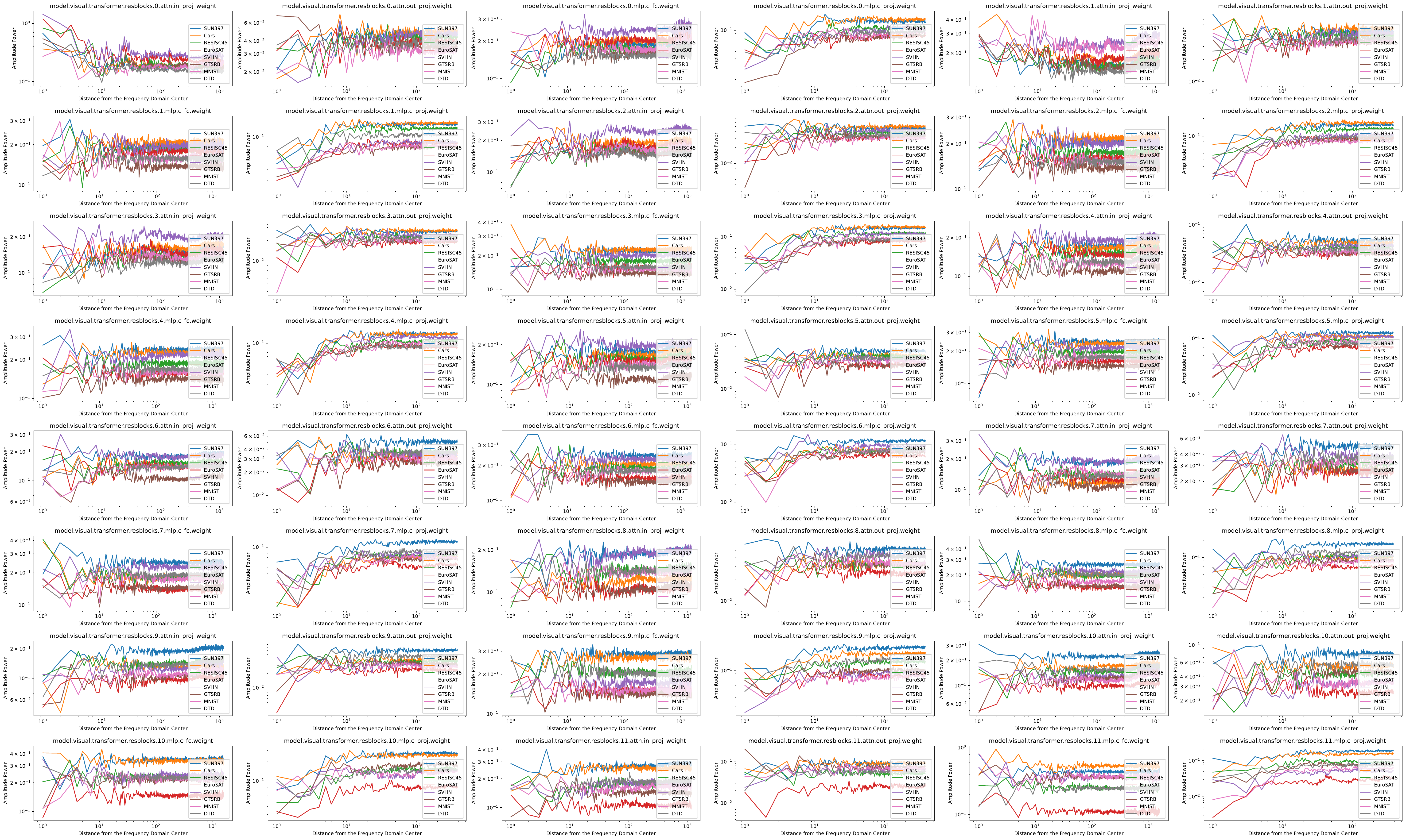} 
\caption{The parameter frequency domain distribution of ViT-B-32 fine-tuning results on eight image tasks.}\label{fig:vit_b_32_full_freq_dist}
\end{figure*}

\begin{figure*}[t] 
\centering  
\includegraphics[width=0.995 \linewidth]{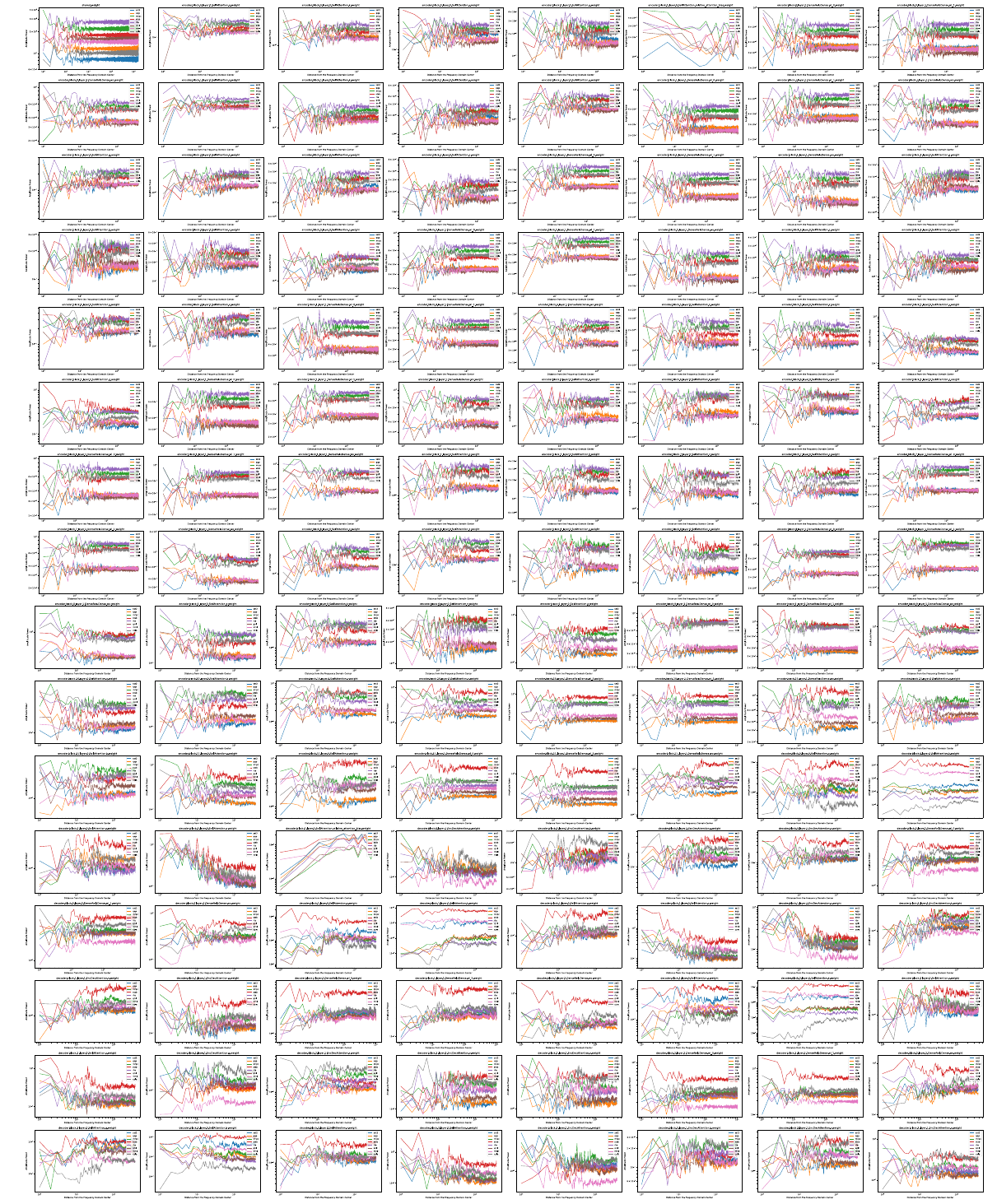} 
\caption{The parameter frequency domain distribution of Flan-T5 fine-tuning results on eight GLUE tasks.}\label{fig:t5_full_freq_dist}
\end{figure*}

Finally, we examine whether low-frequency task interference is widespread across different models. In Fig.~\ref{fig:vit_b_32_full_freq_dist} and Fig.~\ref{fig:t5_full_freq_dist}, we visualize the frequency distribution of fine-tuned parameters for ViT-B/32 and Flan-T5~\cite{chung2024scaling} on their respective tasks. The results show that task interference in the low-frequency region is consistently present across various models. Therefore, our hypothesis is well-supported. Meanwhile, we observe an interesting phenomenon: in some layers, frequency-domain amplitude clustering occurs, meaning that certain models have very similar amplitude distributions. This may be related to the training path during the finetuning process, and we will further investigate this phenomenon in future research. 

\begin{figure*}[t] 
\centering  
\includegraphics[width=1.0 \linewidth]{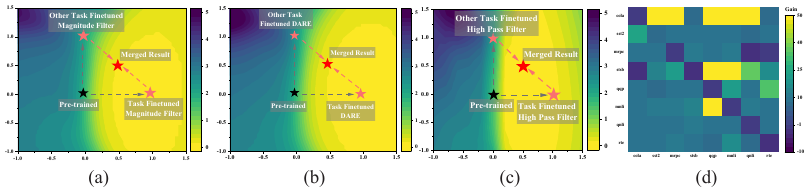} 
\caption{a) Merge by preserving the top 20\% of parameters; b) Merge using the DARE method; c) FR-Merging. \textbf{a-c} complete on ViT/B-32 and test on 8 visual tasks. d) High-pass filtering improves performance over the unfiltered version on RoBERTa and test on GLUE.}\label{fig:rebuttal-fig}
\end{figure*} 

\subsection{Expanded Loss Landscape Analysis}\label{appendix:expanded_frequency_domain_analysis}
In this section, we attempt to analyze why our proposed FR-Merging outperforms existing approaches from the perspective of the loss landscape. A loss landscape comparison with two popular sparse methods, Top-K retention and DARE, is shown in Fig.~\ref{fig:rebuttal-fig} (a-c). Our proposed FR-Merging method brings the two fine-tuned models closer in the loss landscape for the same task, making their linear combination fall closer to the loss basin of the target task, thereby achieving better performance. This aligns with our analysis from the perspective of task interference and is also consistent with the performance improvements in experiments. Therefore, our proposed FR-Merging method offers significant advantages over current approaches.

\subsection{Fourier Transform on Language Models} \label{appendix:fourier_transform_on_language}
In Sec.~\ref{sec:fr-merging}, we analyze the impact of high-pass filtering on the generalization ability of fine-tuned models and their performance on corresponding tasks using ViT-B/32. In this part, we extend the analysis to language models, further demonstrating the effectiveness of high-pass filtering. Fig.~\ref{fig:rebuttal-fig}(d) shows the generalization test of RoBERTa on the GLUE datasets, where filtering out low-frequency information also enhances the generalization ability of fine-tuned language models. This improvement is also reflected in the merging performance improvement for language models. However, compared to the effect on language models, high-pass filtering has a more pronounced impact on ViT. This may be attributed to the nature of image data, which inherently possesses certain high-frequency and low-frequency properties. Nevertheless, in summary, FR-Merging, based on high-pass filtering, is a generalizable, efficient, and high-performing model merging approach.

\subsection{Role of Task Experts}\label{appendix:dis:task experts}
In this section, we demonstrate that introducing additional information can potentially mitigate task conflicts.

We use ViT-B/32 as the pre-trained model and fine-tune it on 8 image classification tasks, following the setup in~\cite{ilharco2022editing}. To prevent the impact of parameter erasure, we select only 5\% of non-overlapping parameters for each task. As a result, after this selection, only task conflicts affect the model merging process. Merging resulted in a significant performance drop, indicating the presence of task conflicts. However, when we double task-specific vectors for each task during inference, performance improves substantially. This suggests that introducing additional information for each task has the potential to resolve task conflicts. 
\begin{table}[t]
    \centering
    \caption{Using ViT-B/32 as the pre-trained model, we fine-tune on eight classification tasks, selecting 5\% non-overlapping parameters per task. Performance improves significantly by adding task-specific parameters during inference.}\label{tab:motivation-overlap}
    \begin{tabular}{ccc}
    \hline
        ~ & Avg acc(\%) & Norm acc(\%) \\ \hline
        Fine-tuned & 90.5 & 100.0 \\ 
        No Overlap 5\% params. & 89.91 & 99.34 \\ \hline
        No Overlap Merging & 67.31 & 74.37 \\
        + Task Knowledge & 84.55 & 93.42 \\ \hline
    \end{tabular}
    
\end{table}

\subsection{Task Experts Construction Analysis}\label{appendix:dis:Approximation_task experts}

\begin{figure}[t] 
\centering  
\includegraphics[width=1.0 \linewidth]{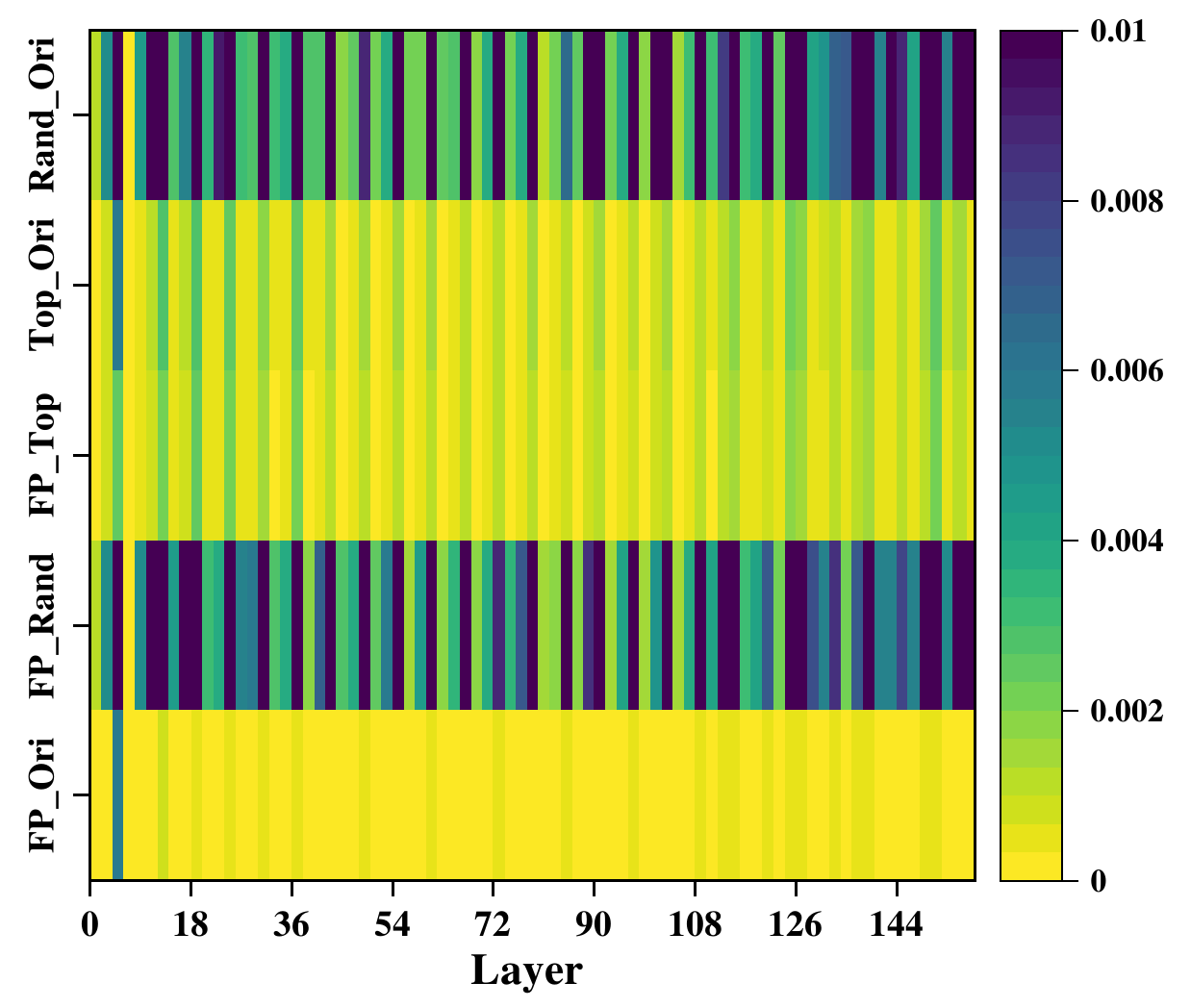} 
\vspace{-0.7cm}
\caption{The analysis evaluates the reasonableness of approximating task-specific information using the most changed parameters during fine-tuning, by calculating the L1 distance of each layer's parameters in CLIP-B/32 fine-tuned on 8 visual tasks with different approximation methods. Here, \textit{Ori} refers to the original parameters, \textit{FP} to low-pass filtering, \textit{Top} to maximum value approximation, and \textit{Rand} to random sampling approximation.} 
\label{figure:expert_approximate.}  
\end{figure} 
In this section, we analyze the construction of experts from two aspects: the parameter selection and rescaling.

\noindent\textbf{Top-K Selection.} We discuss the rationale behind using the parameters with the largest changes during the fine-tuning to approximate task-specific information introduced by fine-tuning, thereby constructing task experts.By comparing L1 distances of model parameters in the frequency domain, with smaller distances indicating better approximation, we assess the effectiveness of different methods.

Based on the analysis in Sec~\ref{sec:expert}, low-frequency information contains more task-specific information but is discarded during model merging in our proposed FR-Merging. Therefore, task experts should compensate for this missing information. First, we briefly explain why approximation methods are preferred over directly saving low-frequency information. If low-frequency data is directly saved, ensuring lightweight storage requires saving frequency-domain data instead of inverse-transformed results. This would necessitate performing inverse transformations during expert invocation, significantly increasing the inference time. Considering the constraints of time and storage in edge-side deployment, a more efficient approximation method is needed.

Next, we compare the extent to which different approximation methods approximate low-frequency information to assess their suitability for constructing task experts. In Fig~\ref{figure:expert_approximate.}, we present the effects of approximating using the top 5\% largest fine-tuned parameters and random sampling. It can be observed that using the parameters with the largest changes during the fine-tuning closely approximates the frequency-domain characteristics of low-frequency information, strongly supporting the rationale for preserving task-specific information by saving the parameters with the largest changes during the fine-tuning. Additionally, the high similarity between low-frequency information and the original information confirms our hypothesis: low-frequency information contains more task-specific information introduced during fine-tuning.

In summary, using the parameters with the largest changes during fine-tuning as task experts to offset task-specific information lost during merging is reasonable. Furthermore, as validated in Sec~\ref{sec:exp:ablation}, employing our proposed rescaling method yields even better results. Thus, the expert construction method proposed in this paper is justified.

\begin{table}[t]
    \centering
    \caption{Ablation of scaling for experts on 8 visual tasks.}\label{tab:rebuttal-ablation-expert-scaling}
    \scalebox{1.0}{
    \begin{tabular}{cccc}
    \hline
        ~ & ViT-B/32 & ViT-B/16 & ViT-L/14 \\ \hline
        w/o scaling & 0.5314 & 0.5323 & 0.6090 \\ 
        w/ scaling & \textbf{0.8484} & \textbf{0.8486} & \textbf{0.8895} \\ \hline
    \end{tabular}
    }
    
\end{table}
\noindent\textbf{Rescaling.} We then discuss the necessity of rescaling. Since we retain only a small portion of the parameters during expert extraction, it alters the mean and variance of parameters, similar to dropout. Our scaling method considers both the mean and the backbone to preserve the similarity between the expert and original model outputs. Table~\ref{tab:rebuttal-ablation-expert-scaling} shows the CKA similarity between expert outputs and original outputs before and after scaling, with higher values indicating better expert performance. Scaling can be found to greatly improve performance compared to not applying it.

\subsection{Role of Router }\label{appendix:dis:Role_of_Router}
In this section, we discuss why a router is necessary for dynamically assigning input data to the appropriate expert. First, let’s review how the router works: when a new data arrives, the router determines which existing expert is best suited to handle it. For example, if we have experts for mathematics and physics, and a math problem appears, the router will automatically route it to the mathematics expert. The mathematics expert will then collaborate with the merged backbone to solve the problem. Manually assigning each input sample to the correct task is impractical, and in real-world applications, we cannot expect users to know in advance which expert is best suited for each piece of data. Therefore, having a router to automate this process is both crucial and aligned with real-world usage scenarios.

\begin{table}[t]
    \centering
    \vspace{-6pt}
    \caption{Ablation study of router training.}\label{rebuttal:router_training}
    \vspace{-2pt}
    \scalebox{0.8}{
    \begin{tabular}{l|cccccc}
    \toprule
        Training Data Size& 0\% & 0.1\% & 0.5\% & 1\% & 2\% & 5\% \\ \midrule 
        ViT-B/32 & 78.1 & 87.0 & 88.2 & 89.7 & 90.0 & 90.1 \\ 
        ViT-L/14 & 88.3 & 92.1 & 93.0 & 93.7 & 93.9 & 94.1 \\ \bottomrule
    \end{tabular}
    }
    \vspace{-8pt}
\end{table}

Meanwhile, in this section, we discuss the amount of data required for training the router. In Tab.~\ref{rebuttal:router_training}, we present the average performance across eight tasks using FREE-Merging with different sizes of training data in the fusion experiments on ViT-B/32 and ViT-B/14. It can be observed that the performance improvement becomes marginal when the training data exceeds 1\%; therefore, in practical applications, we use only 1\% of the data for training. If resources are even more limited, using only 0.1\% of the data can still yield acceptable results.

\subsection{Filtering Frequency Hyperparameter Analysis}\label{appendix:dis:cutoff_hyperparameter}
\begin{table*}[t]
    \centering
    \caption{The performance of Fourier high-pass filtering at different filtering percentages.}\label{tab:appendix:fourier_ratio}
    \scalebox{1.0}{
    \begin{tabular}{l|ccccccccc}
    \hline
        Filtering Percentage & SUN397 & Cars & RESISC45 & EuroSAT & SVHN & GTSRB & MNIST & DTD & Avg. \\ \hline
        Individual & 75.3 & 77.7 & 96.1 & 99.7 & 97.5 & 98.7 & 99.7 & 79.4 & 90.5 \\ \hline
        0\% & 65.3 & 63.4 & 71.4 & 71.7 & 64.2 & 52.8 & 87.5 & 50.1 & 65.8 \\ 
        10\% & 66.1 & 65.3 & 77.2 & 90.0 & 84.3 & 81.0 & 98.2 & 58.4 & 77.6 \\ 
        20\% & 65.6 & 64.5 & 76.9 & 90.3 & 84.6 & 82.6 & 98.5 & 59.1 & 77.7 \\ 
        30\% & 66.2 & 64.5 & 77.2 & 90.1 & 85.4 & 82.3 & 98.5 & 60.0 & 78.1 \\ 
        40\% & 65.2 & 64.4 & 76.1 & 90.4 & 84.5 & 82.2 & 98.5 & 59.0 & 77.5 \\ 
        50\% & 64.9 & 64.4 & 76.3 & 88.9 & 82.3 & 81.2 & 98.4 & 58.9 & 76.9 \\ 
        60\% & 65.4 & 65.0 & 76.9 & 86.7 & 77.3 & 76.6 & 96.9 & 57.7 & 75.3 \\ \hline
    \end{tabular}
    }
    
\end{table*}
\begin{table*}[!ht]
    \centering
    \caption{Results of merging RoBERTa pre-trained models on eight datasets from GLUE benchmark.}\label{tab:lm_medium_all}
    \scalebox{1.0}{
    \begin{tabular}{l|ccccccccc}
    \hline
        Method & CoLA & SST2 & MRPC & STSB & QQP & MNLI & QNLI & RTE & Avg \\ \hline
        Individual & 60.18 & 94.04 & 89.22 & 90.63 & 91.41 & 87.20 & 92.71 & 79.06 & 85.55 \\ \hline
        Weight Averaging & 13.96 & 64.11 & 69.36 & 31.84 & 75.36 & 42.19 & 58.70 & 55.23 & 51.34 \\ 
        
        Task Arithmetic~\cite{ilharco2022editing} & 18.78 & 85.89 & 79.90 & 74.03 & 83.78 & 59.08 & 69.67 & 62.09 & 66.65 \\ 
        Ties-Merging~\cite{yadav2024ties} & 20.48 & 84.40 & 81.13 & 58.19 & 85.70 & 64.65 & 74.81 & 42.96 & 64.04 \\ 
        Breadcrumbs~\cite{davari2023model} & \textbf{25.23} & 85.21 & 78.19 & 68.45 & 81.07 & 64.40 & \textbf{79.83} & 63.18 & 68.19 \\ 
        PCB-Merging~\cite{du2024parameter} & 23.46 & 84.59 & 79.36 & 68.89 & 81.98 & 63.57 & 76.56 & 64.47 & 67.86 \\ 
        FR-Merging(ours) & 15.25 & \textbf{86.81} & \textbf{82.39} & \textbf{80.81} & \textbf{86.21} & \textbf{65.69} & 78.26 & \textbf{64.73} & \textbf{70.02} \\ \hline
        RegMean~\cite{jin2022dataless} & 36.67 & 90.60 & 75.74 & 62.68 & 83.55 & 70.02 & 82.35 & 58.48 & 70.01 \\ 
        EMR-Merging~\cite{huang2024emr} & 37.82 & 87.91 & 82.31 & \textbf{72.90} & 82.61 & 78.91 & 83.24 & 67.92 & 74.20 \\ 
        Twin-Merging~\cite{lu2024twin} & 52.30 & 93.23 & \textbf{88.65} & 68.64 & 81.99 & 78.80 & 88.40 & 74.37 & 78.29 \\ 
        FREE-Merging(ours) & \textbf{54.50} & \textbf{93.69} & 88.46 & 67.04 & \textbf{88.03} & \textbf{80.60} & \textbf{89.90} & \textbf{79.06} & \textbf{80.16} \\ \hline
    \end{tabular}
    }
    
\end{table*}

In this section, we analyze the cutoff frequency of the Fourier high-pass filter. We perform merging using the results after fine-tuning the ViT-B/32 on eight visual classification tasks. We explore the impact of filtering different proportions of low-frequency signals on model merging.

The results are shown in the Table~\ref{tab:appendix:fourier_ratio}. It can be observed that filtering a certain proportion of low-frequency information significantly improves the performance of the model merging. For example, filtering 10\% of the low-frequency information, despite its small proportion, leads to an impressive 12\% improvement in average performance. This is similar to the analysis of low-frequency information in image processing, where low-frequency components contain most of the features in the overall information. In the context of model merging, low-frequency information represents the specialized knowledge brought by fine-tuning. While this specialized information can improve model performance on specific tasks, it severely affects the generalization ability of the model, leading to task conflicts during merging. Therefore, filtering a small proportion of low-frequency information helps minimize model specialization without significantly impacting performance, thus avoiding performance loss caused by task conflicts.

From the experimental results in Table~\ref{tab:appendix:fourier_ratio}, it is evident that our method is robust to the filtering ratio. Even after filtering 40\% of the information, the method still demonstrates strong performance improvements, with a noticeable performance drop only after filtering 60\% of the information. This strongly indicates that the method proposed in this paper is highly robust to the cutoff frequency as a hyper-parameter. The robustness to hyper-parameters makes our method simple to deploy, providing a significant advantage.

\subsection{Fuse to Forget}\label{appendix:dis:Fuse_to_Forget}
In this section, we discuss the forgetting of task-specific components caused by the merging process. In the setting explored in this paper, we aim to preserve the capabilities of multiple individual models simultaneously, so that the merged model can address multiple tasks. Therefore, in our setting, it is crucial to minimize the loss of task-specific capabilities during the fusion process. To address this issue, we propose FR-Merging and FREE-Merging.

On the other hand, \cite{zaman-etal-2024-fuse} suggests that such forgetting is not necessarily harmful, for example, in reducing model bias and other undesirable effects. In their setup, the goal is to actively forget certain undesirable or sensitive information through merging. Thus, the desired behavior of model fusion differs between the two settings.

However, what remains consistent is that the merging process can lead to a loss of each model’s specific capabilities. As discussed earlier, this is primarily due to potential conflicts with the knowledge from other models.

\subsection{Future Works}\label{appendix:future}
In this section, we present the directions for our future work. First, we plan to extend our current merging methods to fine-tuned models with different architectures, thereby enhancing the generalizability and applicability of our approach across a wider range of model types. In addition, we intend to conduct a deeper investigation into the differences between parameter-efficient fine-tuning (PEFT) and full fine-tuning, aiming to understand their respective characteristics and behaviors more thoroughly. Based on these insights, we will develop customized merging strategies specifically tailored to PEFT, enabling more effective integration and deployment of existing PEFT-based models.

\section{Additional Experimental Results}\label{appendix:additional}
\subsection{Merging Vision Models}\label{appendix:additional-30ViTs}
In this section, we comprehensively present the detailed results of model merging using ViT-L/14 as the pre-trained backbone model across all 30 diverse vision datasets introduced and described in Sec.~\ref{sec:exp:vision}, as shown in Table~\ref{tab:30_vit_full}.

From the results, we can see that, consistent with the conclusions presented in the main text, our proposed FR-Merging method significantly outperforms all existing cost-free model merging techniques. It is able to efficiently construct a high-performance merged backbone, demonstrating strong generalization and robustness across various tasks without incurring any additional training or computational overhead. This makes FR-Merging a highly effective solution for model merging in resource-constrained settings. In addition, when we consider the dynamic addition of expert models through FREE-Merging, the method is able to maintain satisfactory performance levels while introducing only a minimal amount of extra computational cost. This flexibility allows for adaptive model expansion depending on the complexity of the task, making FREE-Merging a practical approach for real-world applications where both performance and efficiency are important considerations.

\begin{table*}[t]

\centering
\caption{Task-specific and average performance when merging ViT-B/16 models on \textbf{30} tasks.}
\label{tab:30_vit_full} 
\resizebox{\textwidth}{!}{
\begin{tabular}{l|cccccccccc}
\toprule%

\textbf{Task-specific Acc} & MNIST & Cifar-10 & Vegetables & Food-101 & Kvasir-v2 & Intel-Images & Cars  & EuroSAT & Weather & Cats and Dogs\\ 


\midrule
Individual   & 99.22 & 97.88   & 100.00     & 87.93   & 94.31    & 94.63 & 85.96 & 99.04   & 98.22   & 99.05          \\

\midrule
Weight Averaging   & 27.63 & 42.91   & 83.20      & 68.02   & 25.27    & 82.40 & 7.74  & 24.37   & 61.06   & 91.28     \\  

Task Arithmetic~\cite{ilharco2022editing}& 30.81 & 59.86   & 91.97      & 73.06   & 31.05    & 89.03 & 9.34  & 31.25   & 74.56   & 93.61      \\

Ties-Merging~\cite{yadav2024ties} & 23.21 & 42.82   & 92.31      & 73.22   & 21.09    & 89.39 & 5.30  & 10.98   & 72.86   & 91.88       \\

Breadcrumbs~\cite{davari2023model} & 33.95 & 54.29 & 91.83 & 66.24 & 32.12 & 14.30 & 0.19 & 32.33 & 69.04 & 90.92 \\

FR-Merging(ours) & \textbf{42.78} & \textbf{63.35} & \textbf{93.16} & \textbf{76.25} & \textbf{38.13} & \textbf{90.73} & \textbf{12.00} & \textbf{36.37} & \textbf{75.60} & \textbf{96.69} \\
\midrule

RegMean~\cite{jin2022dataless}     & 90.71 & 89.65   & \textbf{99.10}      & 76.14   & 71.00    & 93.60 & 16.28 & 74.13   & 86.62   & 98.54        \\

AdaMerging~\cite{yangadamerging}  & 81.22 & 87.54   & 97.97      & 75.23   & 22.76    & 91.02 & 0.42  & 44.60   & 89.13   & 96.91        \\

Twin-Merging~\cite{lu2024twin} & 92.10 & 89.34 & 98.54 & 81.43 & 78.45 & 94.14 & 58.12 & 69.43 & 92.23 & \textbf{98.85} \\

EMR-Merging~\cite{huang2024emr} & 93.40 & 90.23 & 98.34 & 83.54 & 76.54 & \textbf{95.12} & 54.34 & 76.64 & 94.54 & 98.65     \\

FREE-Merging(ours) & \textbf{97.30} & \textbf{95.60 }& 99.00 & \textbf{84.65} & \textbf{85.25} & 94.45 & \textbf{58.52} & \textbf{79.12} & \textbf{96.38} & 98.62 \\

\midrule
\midrule
& Dogs  & Fashion & Pet  & LandScape & Flowers & STL-10 & CUB-200-2011 & EMNIST & DTD   & RESISC45 \\
\midrule

Individual   &85.16 & 93.26   & 92.23 & 86.83     & 98.19   & 99.07 & 84.79  & 94.67  & 71.76 & 98.90    \\
\midrule
Weight Averaging & 47.80 & 20.46   & 31.26 & 73.14     & 68.97   & 37.74 & 37.66  & 7.73   & 14.63 & 13.56 \\  

Task Arithmetic~\cite{ilharco2022editing}&47.65 & 37.11   & 33.24 & 79.59     & 80.68   & 39.66 & \textbf{41.86}  & 11.05  & 14.73 & 15.50      \\

Ties-Merging~\cite{yadav2024ties} & 26.03 & 27.05   & 12.84 & 78.27     & 34.33   & 6.17  & 31.28  & 5.61   & 3.71  & 6.79       \\

Breadcrumbs~\cite{davari2023model} & 40.91 & 35.29 & 39.16 & 83.80 & 81.94 & 54.73 & 27.58 & 3.25 & 17.81 & 15.20 \\

FR-Merging(ours) & \textbf{49.86} & \textbf{42.73} & \textbf{41.26} & \textbf{85.80} & \textbf{84.49} & \textbf{59.53} & 41.79 & \textbf{23.80} & \textbf{22.23} & \textbf{19.49} \\

\midrule
RegMean~\cite{jin2022dataless}& 42.89 & 83.42   & 34.62 & 83.64     & 95.26   & 78.94 & 49.78  & 48.67  & 30.53 & 34.66   \\

AdaMerging~\cite{yangadamerging}& 53.09 & 76.76   & 48.34 & 81.98     & 95.69   & 68.91 & 48.19  & 18.02  & 16.68 & 24.83   \\

Twin-Merging~\cite{lu2024twin} & 72.65 & 80.64 & 68.78 & 85.49 & 96.75 & 74.72 & 60.47 & 57.47 & 41.25 & 50.57 \\ 
EMR-Merging~\cite{huang2024emr} & 72.34 & 82.35 & 67.35 & 85.73 & 96.46 & 76.65 & 63.11 & 60.03 & 40.48 & 49.57 \\
FREE-Merging(ours) & \textbf{75.86} & \textbf{86.43} & \textbf{72.57} & \textbf{86.42} & \textbf{97.54} & \textbf{79.01} & \textbf{65.65} & \textbf{63.64} & \textbf{45.71} & \textbf{54.25} \\
\midrule
\midrule

& MangoLeafBD & Beans & Cifar-100 & GTSRB & SVHN& SUN397 & KenyanFood13 & Animal-10N & Garbage & Fruits-360 \\
\midrule
Individual   &100.00     & 97.73 & 89.85    & 95.74 & 96.22 & 78.98  & 85.53  & 92.52      & 93.36                 & 99.63 \\
\midrule

Weight Averaging   &68.58      & 70.98 & 77.98    & 15.00 & 10.88   & 57.42  & 33.55  & 46.00      & 22.89                 & 5.38\\  

Task Arithmetic~\cite{ilharco2022editing}   & \textbf{87.02}      & 84.62 & 80.20    & 37.01 & 17.41  & 55.88  & 36.32  & 51.14      & 25.23                 & 6.15 \\

Ties-Merging~\cite{yadav2024ties}   & 76.58      & 67.22 & 78.61    & 40.74 & 10.54 & 52.69  & 19.90  & 19.13      & 3.91                  & 1.50 \\

Breadcrumbs~\cite{davari2023model} & 79.25 & 85.15 & 75.65 & 38.51 & 22.89 & 47.96 & 26.07 & 17.303 & 28.06 & 1.40 \\

FR-Merging(ours) & 86.50 & \textbf{87.38} & \textbf{82.03} & \textbf{44.19} & \textbf{30.08} & \textbf{56.98} & \textbf{38.90} & \textbf{49.10} & \textbf{32.01} & \textbf{12.00}\\

\midrule

RegMean~\cite{jin2022dataless}   & 98.10      & 92.58 & 82.59    & 56.96 & 66.13  & 58.58  & 57.11  & 68.74      & 65.31                 & 19.79 \\

AdaMerging~\cite{yangadamerging}     & 99.13      & 93.38 & 84.19    & 59.90 & 25.70   & 64.09  & 48.66  & 66.55      & 38.54                 & 7.94  \\
Twin-Merging~\cite{lu2024twin} & 100.0 & 94.24 & 85.57 & 65.87 & 75.49 & 71.25 & 63.81 & 71.67 & 70.82 & 25.56 \\
EMR-Merging~\cite{huang2024emr} & 100.0 & 95.45 & 86.43 & 65.94 & 76.46 & 72.45 & 65.43 & 73.84 & 72.24 & 28.24 \\ 

FREE-Merging(ours) & \textbf{100.0} & \textbf{96.40} & \textbf{88.65} & \textbf{68.63} & \textbf{82.23} & \textbf{76.78} & \textbf{69.45} & \textbf{78.40} & \textbf{75.84} & \textbf{37.97} \\

\midrule

\midrule
\textbf{Average Acc}& \multicolumn{2}{c|}{Individual}& \multicolumn{2}{|c|}{Weight Averaging}&
\multicolumn{2}{c|}{Task Arithmetic}&
\multicolumn{1}{c|}{Ties-Merging}&
\multicolumn{1}{c|}{Breadcrumbs}&
\multicolumn{2}{c}{\textbf{FR-Merging(ours)}}
\\
\midrule
Acc& 
\multicolumn{2}{c|}{93.02}&
\multicolumn{2}{c|}{42.52}&
\multicolumn{2}{c|}{48.89}&
\multicolumn{1}{c|}{37.53}&
\multicolumn{1}{c|}{43.57}&
\multicolumn{2}{c}{\textbf{53.91}}
\\
\bottomrule

\midrule
\textbf{Average Acc}& 
\multicolumn{2}{c|}{Individual}& 
\multicolumn{1}{c|}{RegMean}&
\multicolumn{1}{c|}{AdaMerging}&
\multicolumn{2}{c|}{Twin-Merging}&
\multicolumn{2}{c|}{EMR-Merging}&
\multicolumn{2}{c}{\textbf{FREE-Merging(ours)}}
\\
\midrule
Acc& 
\multicolumn{2}{c|}{93.02}&
\multicolumn{1}{c|}{68.14}&
\multicolumn{1}{c|}{60.25}&
\multicolumn{2}{c|}{75.52}&
\multicolumn{2}{c|}{76.39} &
\multicolumn{2}{c}{\textbf{79.67}} 
\\
\bottomrule

\end{tabular}
}

\end{table*}

\begin{table*}[!ht]
    \centering
    \vspace{25pt}
    \caption{Results of merging $(IA)^3$ models with T0-3B as pre-trained model on eleven NLP tasks.}\label{tab:lm_ia3_all}
    \scalebox{0.77}{
    \begin{tabular}{l|cccccccccccc}
    \hline
        Method & RTE  & CB  & Winogrande  & WiC  & WSC  & COPA  & H-SWAG  & Story Cloze  & ANLI-R1  & ANLI-R2  & ANLI-R3  & Avg. Acc \\ \hline
        Individual & 82.7 & 95.8 & 75.1 & 71.7 & 65.3 & 85.3 & 44.4 & 94.9 & 70.2 & 46.5 & 53 & 71.4 \\ 
        Traditional MTL & 88.6 & 95.8 & 75.5 & 61.1 & 80.6 & 94.1 & 42.3 & 97.6 & 70.5 & 49.8 & 47.7 & 73.1 \\ \hline
        Fisher Merging~\cite{matena2022merging} & 83.3 & 83.3 & 56.7 & 54.2 & 58.3 & 83.1 & 42.2 & 94.1 & 45.9 & 41.0 & 42.2 & 62.2 \\ 
        
        Task Arithmetic~\cite{ilharco2022editing} & 74.1 & 83.3 & 62.8 & 49.1 & 49.3 & 87.5 & 41.5 & 95.3 & 60.8 & 49.4 & 50.0 & 63.9 \\ 
        Weight Averaging & 81.2 & 58.3 & 53.8 & 55.2 & 53.5 & 80.9 & 40.1 & 92.5 & 43.3 & 39.2 & 40.2 & 58.0 \\ 
        Ties-Merging~\cite{yadav2024ties} & 78.0 & 83.3 & 67.9 & 57.6 & 59.7 & 81.7 & 42.8 & 90.3 & \textbf{66.9} & \textbf{51.3} & \textbf{51.1} & 66.4 \\ 
        Breadcrumbs~\cite{davari2023model} & 71.9 & 59.4 & 53.1 & 31.2 & 64.1 & 78.1 & 46.9 & 90.2 & 50.0 & 31.2 & 25.0 & 54.6 \\ 
        PCB-Merging~\cite{du2024parameter} & \textbf{85.9} & 83.3 & 61.9 & 57.1 & 63.9 & 82.4 & 42.7 & 91.2 & 64.2 & 47.8 & 45.9 & 66.1 \\ 
        FR-Merging(ours) & 81.3 & \textbf{83.8} & \textbf{68.3} & \textbf{57.9} & \textbf{65.2} & \textbf{88.4} & \textbf{48.6} & \textbf{93.5} & 56.2 & 46.2 & 46.3 & \textbf{66.9} \\ \hline
        RegMean~\cite{jin2022dataless} & 81.2 & 58.3 & 53.8 & 55.2 & 53.5 & 80.9 & 40.1 & 92.5 & 43.3 & 39.2 & 40.2 & 58.0 \\ 
        EMR-Merging~\cite{huang2024emr} & 81.8 & 87.5 & 66.6 & 56.1 & 65.3 & 82.4 & 44.7 & 93.6 & 65.7 & 43.8 & 50.8 & 67.1 \\ 
        Twin-Merging~\cite{lu2024twin} & 81.2 & 85.6 & 65.4 & 57.2 & \textbf{66.3} & 81.6 & 44.4 & 92.9 & 66.5 & 42.4 & 51.2 & 66.8 \\ 
        FREE-Merging(ours) & \textbf{83.2} & \textbf{89.3} & \textbf{69.9} & \textbf{58.4} & 65.5 & \textbf{84.3} & \textbf{45.2} & \textbf{94.2} & \textbf{67.9} & \textbf{45.2} & \textbf{52.4} & \textbf{68.7} \\ \hline
    \end{tabular}
    }
    
\end{table*}

\subsection{Merging Language Models}\label{appendix:merging_language_models}
\begin{table*}[!ht]
    \centering
    \caption{Results of merging LoRA models with QWEN-14B as pre-trained model on three generative tasks.}\label{tab:lm_qwen_lora_all}
    \scalebox{1.0}{
    \begin{tabular}{l|cccc}
    \toprule
        Method & MMLU & TruthfulQA & BBQ & Avg. \\ \midrule
        Pre-trained & 69.30 & 51.27 & 80.69 & 67.09 \\ 
        Fine-tuned & 68.35 & 54.34 & 93.53 & 72.07 \\ \midrule
        Weight Averaging & 68.10 & 50.01 & 82.31 & 66.80 \\ 
        Task Arithmetic~\cite{ilharco2022editing} & 67.62 & 53.38 & 78.24 & 66.41 \\ 
        Task Arithmetic (w/ DARE)~\cite{yu2024language} & 67.82 & 52.66 & 82.83 & 67.77 \\ 
        Ties-Merging~\cite{yadav2024ties} & 68.27 & 50.01 & \textbf{84.10} & 67.46 \\ 
        Ties-Merging (w/ DARE)~\cite{yu2024language} & \textbf{69.32} & 53.07 & 81.19 & 67.86 \\ 
        Breadcrumbs~\cite{davari2023model} & 68.24 & 52.88 & 79.20 & 66.77 \\ 
        PCB-Merging~\cite{du2024parameter} & 67.23 & 52.48 & 80.37 & 66.69 \\ 
        FR-Merging(ours) & 68.16 & \textbf{53.39} & 82.44 & \textbf{68.00} \\ \midrule
        RegMean~\cite{jin2022dataless} & 67.89 & 52.45 & 82.24 & 67.52 \\ 
        EMR-Merging~\cite{huang2024emr} & 67.82 & 55.01 & 90.13 & 70.98\\ 
        Twin-Merging~\cite{lu2024twin} & 68.32 & 55.76 & 90.98 & 71.68 \\ 
        FREE-Merging(ours) & \textbf{68.83} & \textbf{57.39} & \textbf{92.14} & \textbf{72.78} \\ \bottomrule
    \end{tabular}
    }
    
\end{table*}
In this section, we present the complete and detailed experimental results of merging a diverse set of language models, each fine-tuned under different settings, across multiple widely-used benchmark datasets that cover both classification and generative tasks.

First, we present the merging results using RoBERTa as the pre-trained model, which was fine-tuned individually on the eight tasks included in the GLUE benchmark. These results are summarized in Table~\ref{tab:lm_medium_all}. Our proposed method achieves the best performance on nearly all of the datasets, highlighting its strong generalization capability when applied to language models. In addition, Table~\ref{tab:lm_ia3_all} reports the merging results across 11 language classification tasks, where T0-3B serves as the pre-trained model and $IA^3$ is used for fine-tuning. Our method also demonstrates a clear advantage in this setting. Finally, Table~\ref{tab:lm_qwen_lora_all} provides the results for merging with QWEN-14B as the pre-trained model and LoRA as the fine-tuning technique, evaluated on three generative tasks. In all these experiments, both FR-Merging and FREE-Merging deliver consistently superior performance compared to other existing model merging methods, confirming their effectiveness across various fine-tuning strategies and model types.

Overall, our proposed methods demonstrate strong generalization capabilities when applied to language models, consistently delivering robust performance across a wide range of tasks, architectures, and fine-tuning strategies.


\end{document}